\documentclass[runningheads]{llncs}

\usepackage[mobile]{eccv}
\usepackage{eccvabbrv}
\renewcommand{\eg}{{e.g}\onedot} 
\renewcommand{\ie}{{i.e}\onedot}
\usepackage{graphicx}
\usepackage{booktabs}
\usepackage{multirow}
\usepackage{pifont}
\usepackage[accsupp]{axessibility}
\usepackage{orcidlink}
\usepackage{booktabs}
\usepackage{float}
\usepackage{amsfonts} 
\usepackage{lineno}
\usepackage{overpic}
\usepackage{bm}
\usepackage{xcolor}

\newcommand{\CE}{\text{CE}}
\newcommand{\IoU}{\text{IoU}}
\newcommand{\MSE}{\text{MSE}}

\newcommand{\ioul}{\IoU\textsubscript{learn}}
\newcommand{\ioua}{\IoU\textsubscript{align}}
\newcommand{\ioup}{\IoU\textsubscript{seg}}

\newcommand{\biasdata}{$\mathcal{D}_{\text{bias}}$}
\newcommand{\unidata}{$\mathcal{D}_{\text{uni}}$}
\newcommand{\rebodata}{$\mathcal{D}_{\text{ReBO}}$}

\newcommand{\Atheta}{\text{$\mathcal{A}_\theta$}}
\newcommand{\Athetai}{\text{$\mathcal{A}^{-1}_\theta$}}
\newcommand{\Athetap}{\text{$\hat{\mathcal{A}_{\theta}}$}}
\renewcommand{\Athetap}{\text{${\mathcal{A}_{\hat{\theta}}}$}}
\newcommand{\Athetari}{\text{$\mathcal{A}_{\theta_2}^{-1}$}}
\newcommand{\Athetarp}{\text{$\hat{\mathcal{A}}_{\theta_2}$}}
\renewcommand{\Athetarp}{\text{${\mathcal{A}}_{\hat{\theta}_2}$}}
\newcommand{\Athetar}{\text{$\mathcal{A}_{\theta_2}$}}

\newcommand{\x}{\ensuremath{I_x}}
\newcommand{\dom}{\mathbb{P}^2}
\newcommand{\y}{\ensuremath{I_y}}
\newcommand{\yprime}{\ensuremath{{I_{y'}}}}
\newcommand{\yprimeTwo}{\ensuremath{{I_{y'_2}}}}
\newcommand{\yhat}{\ensuremath{{I_{\hat{y}}}}}
\newcommand{\yhata}{\ensuremath{{I_{\hat{y}_a}}}}

\usepackage{eso-pic}
\usepackage{xcolor}

\AddToShipoutPictureFG*{%
  \AtPageLowerLeft{%
    \ifnum\value{page}=1
      \put(34.1,15){%
        \parbox{1.1\textwidth}{%
          \small\color{gray}
          Published in
          \emph{European Conference on Computer Vision (ECCV)}, 2026.
        }%
      }
    \fi
  }
}
\begin{document}

\title{Align and Segment: Unsupervised Learning for Building Segmentation From Misaligned Labels} 
\titlerunning{Align and Segment}

\author{Venkanna Babu Guthula\inst{1}\orcidlink{0000-0001-5902-5905} \and
Oswin Krause\inst{1}\orcidlink{0000-0002-0990-559X} \and
Dimitri Gominski\inst{1}\orcidlink{0000-0002-8135-1341} \and \\
Hui Zhang\inst{1}\orcidlink{0000-0002-0992-5830} \and
Johan Mottelson\inst{2}\orcidlink{0000-0002-3440-288X} \and
Ankit Kariryaa\inst{1}\orcidlink{0000-0001-9284-7847} \and
Nico Lang\inst{1}\orcidlink{0000-0001-8434-027X} \and
Christian Igel\inst{1}\orcidlink{0000-0003-2868-0856}}
\authorrunning{Guthula et al.}

\institute{University of Copenhagen, Copenhagen, Denmark \and
Royal Danish Academy, Copenhagen, Denmark \\
\email{\{vegu, igel\}@di.ku.dk}}

\maketitle

\begin{abstract}

  Supervised learning for image segmentation typically requires spatially aligned image and label sets. When images and labels originate from different sources, the pairing may be misaligned, which can significantly deteriorate the performance of the learned models. This is especially common in remote sensing, when aerial or satellite images are co-registered with labels from another source (\eg, OpenStreetMap). In this work, we propose a novel approach for training on misaligned labels, where we simultaneously learn the label alignment. Our align and segment (AnS) approach builds on the spatial transformer module to transform the misaligned labels using an affine transformation to provide a better learning target for a canonical semantic segmentation network. We prevent shortcut learning of misaligned labels in these semantic segmentation networks through a self-supervised regularization loss and show that it is complementary to data augmentation, especially for systematically misaligned training data. A decisive characteristic of our AnS approach is that it learns \emph{without requiring any ``golden'' labels}. We experimentally show on both synthetic and real-world data from different cities that our approach enables high-quality building segmentation and precise label-image alignment at the same time. Code and derived datasets are available at \url{https://github.com/venkanna37/align-and-segment}.

\begin{figure}[t]
    \centering
\begin{minipage}[b]{0.5\textwidth}
    \centering
    \includegraphics[width=\linewidth]{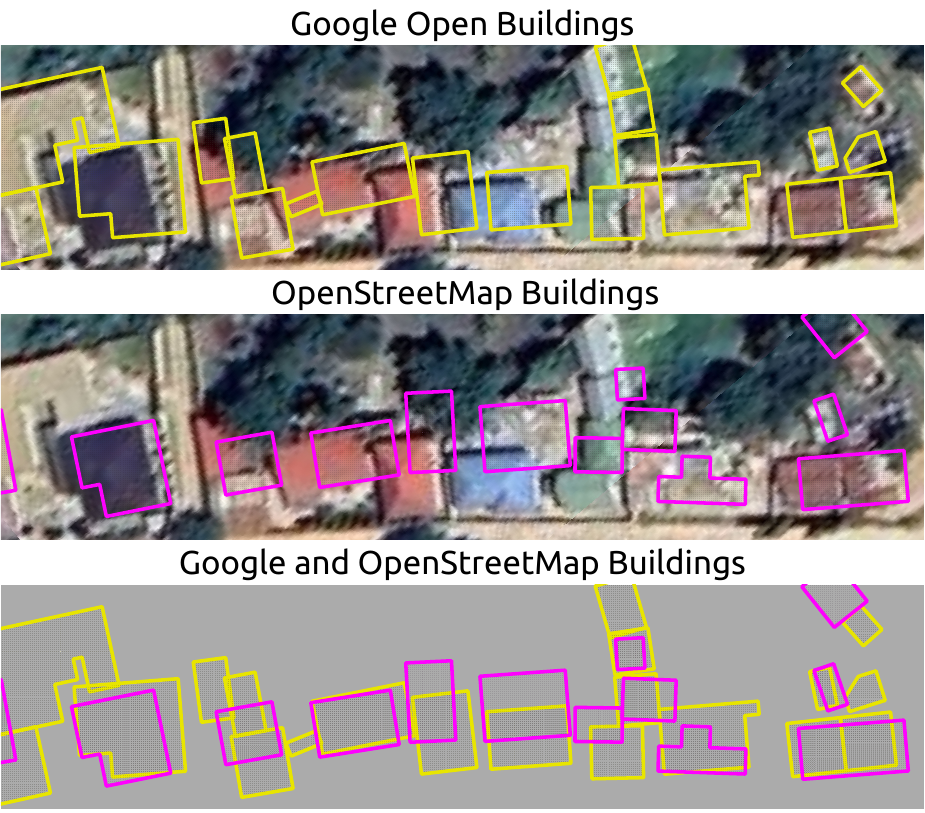}
\end{minipage}
\hfill
\begin{minipage}[b]{0.48\textwidth}
    \centering
    \includegraphics[width=\linewidth, trim=0 0.6em 0 0]{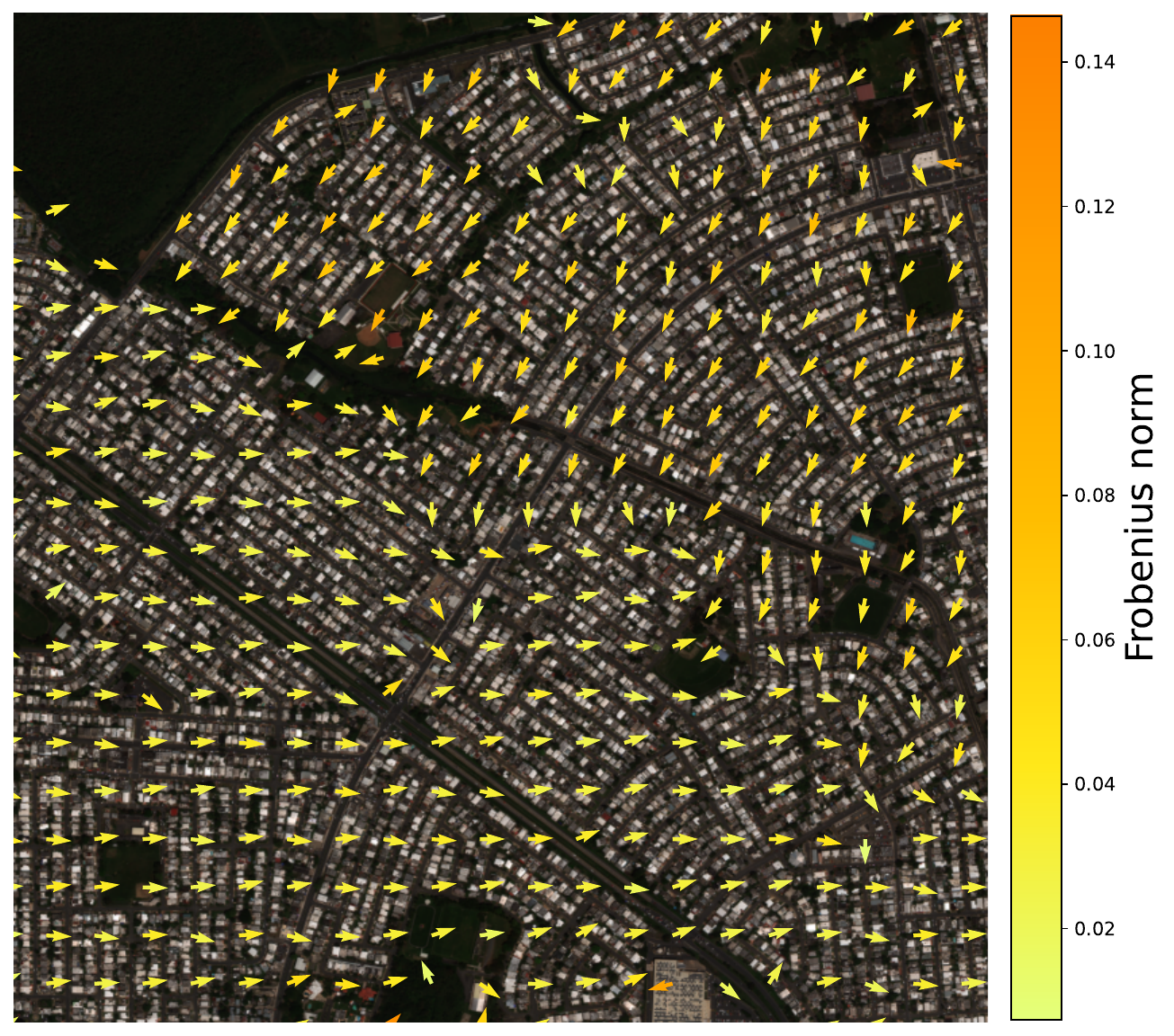}
\end{minipage}
    \caption{\textbf{Misalignment of building footprints with independent imagery.} Data \eg from Google Open Buildings~\cite{sirko2021continental} or OpenStreetMap (OSM)~\cite{OpenStreetMap} neither align with independent satellite imagery nor with each other. The right figure shows our estimated misalignment of OSM data, highlighting the locally systematic label noise. The arrows indicate estimated translations, with color representing the displacement magnitude.}
    \label{fig:teaser}
\end{figure}

\end{abstract}

\section{Introduction}
\label{sec:intro}
Label noise is a ubiquitous problem in computer vision. In semantic segmentation such noise can either be caused by imprecise annotations or by misalignments when data from different sources are combined.  
When training a segmentation network on such noisy labels, it is prone to predict the noisy or misaligned labels given enough capacity. The network's performance can sharply deteriorate if the noise is too high~\cite{maiti2022effect}. This issue is even worse when the model is trained on systematic label noise, for example, on biased misalignments.

In remote sensing, where images (\eg, satellite or drone imagery) and labels (\eg, crowd-sourced or field-collected) often originate from different sources, label noise is particularly common, especially spatial misalignment of images and label masks. For example, overlaying OpenStreetMap (OSM)~\cite{OpenStreetMap} building footprints against other public building maps reveals \emph{large and systematic} local transformations of several pixels (\cref{fig:teaser}). 
This bottleneck has led to an under-utilization of large swaths of hand-curated data such as OSM for training deep learning models at scale and in regions where high-quality building data are scarce.

At first glance, jointly modeling label alignment and segmentation may appear to be a straightforward solution. However, such a disentanglement is nontrivial and a naive implementation in two sub-networks is prone to fail, due to the ability of segmentation networks to simply learn the misaligned labels. An alternative popular approach is to view the problem as a multimodal image alignment task and to estimate the transformation between the image and the label maps using feature maps learned for the different modalities \cite{Zampieri_2018_ECCV, Girard_2019, Girard2019_ACCV}. 
However, these approaches usually assume that the observed misalignments in the dataset are small and unbiased. In remote sensing these assumptions often do not hold, as can be seen in \cref{fig:teaser}, where large and biased misalignments of images can be introduced by erroneous orthorectification (\ie, projecting images onto terrain) and other satellite image preprocessing steps as well as errors made in the land registry offices.

In this work, we propose a learning-based approach for dealing with misaligned labels in segmentation without requiring any ``golden'' (clean, ground-truth) labels. Our solution involves several components of regularization, including a new loss that allows two sub-networks to learn label alignment and segmentation respectively. 
We focus on the segmentation of building footprints from high-resolution satellite imagery, an essential resource for applications such as urban planning as well as disease and disaster risk assessment in informal settlements \cite{bettencourt_infrastructure_2025}. Especially for the latter, curated data is rare, as benchmark datasets overwhelmingly focus on urban areas in Europe and the USA \cite{maggiori2017can,bradbury2016aerial}. While open label maps, such as OSM polygons, exist for these regions, they require extensive alignment and correction by human experts for reliable training~\cite{guthula2025drone}, which limits the scalability of this valuable label source.

The main contribution of this work is a method for unsupervised learning of multi-modal alignment between image modalities and segmentation maps, without the need for \emph{any} golden labels and without assuming unbiased transformations between the modalities. We summarize our contributions as follows:

\begin{enumerate}
    \item We propose \emph{align and segment (AnS)} as a methodology that can be combined with any existing semantic segmentation network to directly learn from misaligned labels \emph{without using any golden labels}. For this, we propose a combination of a self-consistency loss and data augmentations to directly mitigate the learning of misaligned segmentation maps and biased transformations.
    
    \item We empirically show that AnS applied to building footprint alignment generalizes across different cities and can learn alignment and segmentation for a large range of noise levels and diverse label characteristics, such as building sizes and densities.
    \item We compared our method with multiple baselines of supervised and unsupervised methods, and evaluated on real-world and synthetic datasets.
\end{enumerate}

\section{Related work}

\textbf{Label noise} is a common issue in computer vision, and has become topical in the context of deep learning where models are trained on databases of increasing scale but decreasing curation efforts. For image classification, popular research directions include noise-robust loss functions \cite{zhang_generalized_2018}, tailored regularization \cite{liu_early-learning_2020} and label correction \cite{kun_probabilistic_2019}. Extending these ideas to semantic segmentation is straight-forward \cite{zhang_characterizing_2020, zhu_pick-and-learn_2019}, but under the unrealistic assumption that label noise is independent and identically distributed among pixels \cite{yao_learning_2023}. Accordingly, recent work has tackled label noise in semantic segmentation specifically, with noise modeling \cite{yao_learning_2023} or by exploiting the phenomenon of noise robustness in early learning stages \cite{liu_adaptive_2022}.

\textbf{Misaligned supervision} tasks are recurrent in remote-sensing and medical imaging. A common approach is to solve the misalignment problem independently of the segmentation problem, by first aligning the noisy segmentation and image~\cite{Zampieri_2018_ECCV,Girard_2019, Girard2019_ACCV, chen_autocorrect_2019}. Most approaches require at least a small amount of high-quality annotations~\cite{Zampieri_2018_ECCV, zorzi2020map, simone2020learning}. Closest to our work is~\cite{chen_autocorrect_2019}, which introduces a consistency loss that applies pairs of random transformations to the misaligned labels and penalizes the network predictions when they are inconsistent with respect to the applied transformations. However, without additional labels, the neural network can predict any biased set of predictions as long as they are relatively consistent with each other (e.g., all predictions may be erroneously shifted by the same constant offset). Our approach improves on this, as we predict transformations based on pairs of segmentation maps, which allows us to enforce absolute consistency with respect to a known random transformation. In~\cite{Girard_2019}, a multi-round procedure is proposed where, in each round, a model is trained to predict random transformations applied to the misaligned segmentation maps. This model is used to align the segmentation maps, which are then used to re-train the model in the next round. This method cannot work in the presence of biased transformations as the network is trained to predict an unbiased distribution of transformations. Although several studies also consider learning the segmentation map~\cite{simone2020learning, li2025dragosm, Girard_2019}, none of them have attempted to jointly learn alignment and segmentation with the amount of shifts commonly found in remote sensing data. However, small translation and rotation errors were addressed by \cite{simone2020learning, Girard_2019, zorzi2020map}.

\section{Methodology}

\begin{figure*}[tb]
  \centering
    \includegraphics[width=\textwidth]{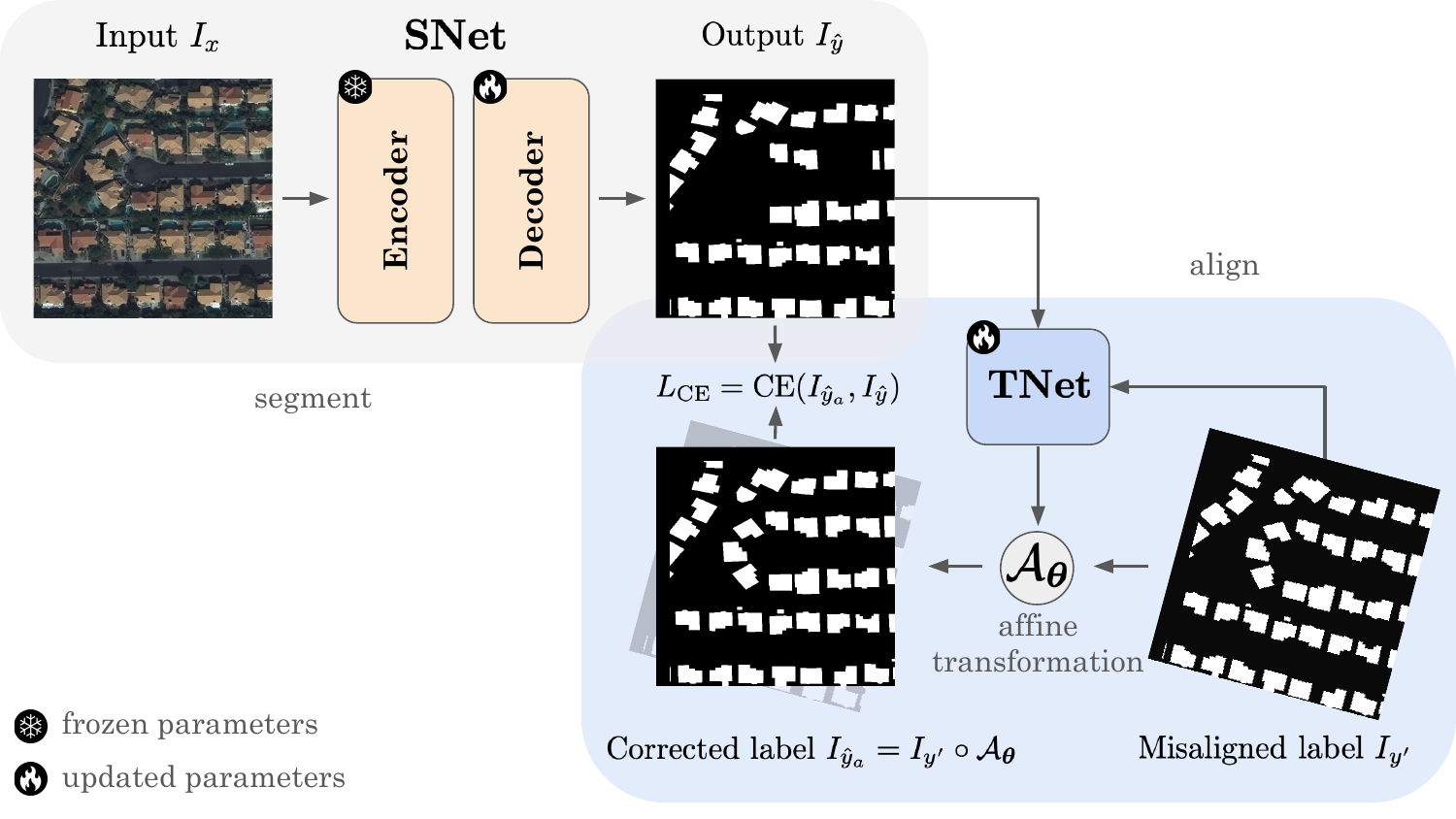}
  \caption{\textbf{Illustration of our align and segment (AnS) methodology.} 
  To directly learn semantic segmentation from misaligned labels, we introduce TNet, a lightweight module that learns to correct misaligned label maps while learning segmentation. This model-agnostic approach can be integrated in a plug-and-play manner with any segmentation network (SNet), as TNet directly operates on semantic segmentation maps.
  }
  \label{fig:final_method}
\end{figure*}
We assume in the following that images and segmentation masks are defined in homogeneous coordinates and w.l.o.g.{} consider single channel images and binary segmentation problems, that is, we consider images as continuous functions of the form $I:\dom \to \mathbb{R}$.

\subsection{Model architecture}

In our framework, the inputs are given by an input image \x{} and a corresponding misaligned segmentation mask \yprime. We assume that $\yprime$ is related to the aligned (unknown) segmentation mask \y{} by an affine transformation $\mathcal{A}_{\theta^*}$ of the form
\begin{equation}
    \Atheta = \begin{bmatrix}
        \cos(\alpha) & -\sin(\alpha) & t_x \\
        \sin(\alpha) & \cos(\alpha) & t_y \\
        0 & 0 & 1   
    \end{bmatrix}\enspace, \label{eq:affine}
\end{equation}
where $ \theta=(\alpha, t_x, t_y)$ are the three degrees of freedom of the transformation. With this, we have $\yprime=I_y \circ \mathcal{A}_{\theta^*}$.
To predict this transformation, we employ a pair of networks, a TNet to predict the transformation between the image and the misaligned label mask, and an SNet to predict the semantic segmentation map from the image, see Fig.~\ref{fig:final_method}. 

\paragraph {SNet.} The segmentation network is a canonical deep neural network for semantic segmentation, mapping input images $\x:\dom\to\mathbb{R}$ to probabilistic segmentation masks $\yhat:\dom\to[0,1]$.

\paragraph {TNet.} For the transformation network, our approach is inspired by the Spatial Transformer Network (STN, \cite{NIPS2015_33ceb07b}) architecture introduced for classification\footnote{The name STN is \emph{not} related to transformer networks based on attention~\cite{vaswani2017attention}.}, see the lower part of Fig.~\ref{fig:final_method}. In our AnS approach, the TNet takes as input the misaligned segmentation mask
$\yprime$ and the segmentation mask $\yhat$, predicted by the SNet. Given  $\yprime$ and $\yhat$, the TNet predicts the parameters of the affine transformation $\Athetap$. This predicted affine matrix is applied to $\yprime$ to yield the realigned segmentation mask $\yhata=\yprime \circ \Athetap=\y \circ \mathcal{A}_{\theta^*} \circ \Athetap$. 

To constrain the possible transformations, we apply an activation function $a(x)=c\tanh(x)$ to the network outputs.
The scaling factor $c$ limits the maximum alignment we can predict with the TNet, and can be based on the expected misalignment. 
The architecture can be trained via the cross entropy-loss 
\begin{equation}
    L_{\CE} = \CE(\yprime\circ\Athetap, \yhat) =  \CE(\yhata, \yhat )\enspace.
\end{equation}
In practice, when dealing with rasterized rectangular images, the transformation \Athetap{} leads to non-overlapping areas at the borders, which are excluded from the  loss computation. 

\subsection{Handling transformation bias}
 The naive approach of minimizing $L_{\CE}$ fails, as the label map $\yhat$ predicted by the SNet can itself be translated with respect to \x. Counterintuitively, this is the expected outcome. At the start of optimization, the TNet cannot predict the correct $\Athetap\approx A_{\theta^*}^{-1}$, and will likely predict a transformation close to the identity. Thus, at the start of optimization, we have
$\yhat\approx \yprime$ and thus, the loss of the SNet is given by comparing its prediction to the unaligned annotations. As a result, in the presence of bias in the label alignments, the SNet is driven towards predictions that are transformed to compensate for the misalignment.
Since the cross-entropy can be potentially minimized to zero for the transformed maps, this is a stable solution, and thus 
the TNet will not learn the underlying translation.

We therefore developed two countermeasures: first, a consistency loss is added as an additional regularizer, which enables the TNet to move away from this unwanted local optimum. Second, data augmentations make it more difficult for the SNet to learn the biases in the dataset.

\subsubsection{Consistency loss.}  To compute the consistency loss, we draw a random transformation $\Athetar$ and apply it to the misaligned label map $\yprime$ to obtain $\yprime \circ \Athetar$ similar to AutoCorrect~\cite{chen_autocorrect_2019}. Then we apply the TNet again to the input pair $(\yprime \circ \Athetar,\yprime)$ and predict $\Athetarp$, which should ideally fulfill $\Athetarp\approx \Athetari$. Since in this case, we know the optimal output of the TNet, we can measure both the cross-entropy of the re-aligned label-maps as well as the distance between $\Athetarp$ and $\Athetari$. Most importantly, this prediction is independent of the segmentation map predicted by the SNet, allowing the TNet to move away from the unwanted local optimum. Moreover, it provides a learning signal early during training when the SNet has not developed a proper segmentation map yet. 

The deviation of the predicted $\Athetarp$ and the optimal $\Athetari$ is measured by the
mean squared error (MSE) between the transformation matrix elements and the intersection over union (IoU) between the rasterized $\yprimeTwo\circ\Athetarp$ and $\yprime$: 
\begin{equation}\label{eq:loss2}
    L_{\text{Con}} = \lambda \MSE(\Athetari, \Athetarp) + \IoU(\yprime,  \yprime \circ (\Athetar \circ \Athetarp))
\end{equation}
The IoU loss emphasizes the overlapping regions between the two building masks, providing strong gradients when there is some overlap, while the MSE loss complements it by ensuring gradient flow even when there is little or no overlap between the two masks given to the TNet.
As for the cross-entropy loss, non-overlapping regions are not considered in the $\IoU$ computation.

\subsubsection{Augmentation.}
Even with the above loss functions, there is a risk that the segmentation model remains misaligned, as the required changes in the model parameters can be too large and the optimisation can get trapped in a local optimum.
To overcome this issue, we applied strong augmentations both to the input image and misaligned mask to counteract the bias in the dataset. We do this by applying simple geometric augmentations such as vertical and horizontal flips, and rotations ($90^\circ$, $180^\circ$ and $270^\circ$) to the image pairs. For example, a horizontal flip of an image pair with an existing misalignment of 50 pixel translation on the $x$-axis changes to -50 pixel translation on the $x$-axis on the flipped image. In expectation, this removes the translation bias in the dataset.

\subsubsection{Evaluation.}
We evaluate the performance using two variants of Intersection over Union (IoU) using `golden' semantic reference maps $\y$ that are aligned with the input image $\x$.
Firstly, \ioup{} measures the performance of the segmentation network by comparing $\yhat$ and $\y$:
\begin{equation}
    \ioup{} = \IoU(\yhat, \y)
\end{equation}
Secondly, \ioua{} measures the performance of the alignment network.
It transforms the reference mask with a known transformation $\Athetai$ and measures how well the TNet can learn the inverse transformation:
\begin{equation}
    \ioua{} = \IoU(\y, \y\circ\Athetai \circ \Athetap )
\end{equation}
Errors in \ioua{} indicate erroneous estimates of the transformation parameters given a perfect segmentation.

Lastly, since there is a risk of the segmentation model learning misaligned labels directly, we are interested in evaluating if the modules learn the expected behavior, \ie separating the alignment and segmentation. 
Therefore, we also report \ioul{} computed by comparing $\yhata$ and $\yhat$:
\begin{equation}
        \ioul{} = \IoU(\yprime\circ \Athetap , \yhat ) 
                =  \IoU(\yhata, \yhat )
\end{equation}
A high \ioul{} but low \ioup{} and/or low \ioua{} indicates that the segmentation model learned to predict the misaligned labels and that the disentanglement  of alignment and segmentation failed.

\section{Dataset preparation}
We studied building segmentation in remote sensing imagery from various cities across the world (see \cref{tab:dataset_overview}).
To quantitatively study both random and systematic misalignment, 
we considered labeled data from Las Vegas, Paris, and Khartoum. We
created two datasets per city from SpaceNet 2~\cite{van2018spacenet}, which contains WorldView-3 satellite imagery with 30~cm ground sampling distance and respective building footprints. The first is the biased dataset, \biasdata, where all masks are systematically translated by 50 pixels along the $x$-direction.
As demonstrated in \cref{fig:teaser}, label misalignment in geospatial applications is dominated by systematic noise. Learning from systematically misaligned data 
is particularly difficult, as a segmentation model can simply learn this misalignment, and, as a result,  the input and the output map remain (geospatially) misaligned.
The second dataset, \unidata, was generated using randomly sampled affine transformations.
If not further specified, we sampled transformations within the range $[-50, 50]$ pixels for $t_x$ and $t_y$ and $[-4.5, 4.5]$ degrees of $\alpha$. This dataset spans a wide range of misalignment levels, from minimal 
to large geometric deviations. Correcting misalignment from \unidata{} is also challenging when the model is not sensitive to small misalignments.
Fig.~\ref{fig:qualitative_predictions} shows example images 
from both datasets.

\begin{table}
  \centering
  \caption{\textbf{Dataset overview.} 
  Our synthetic datasets were constructed from the labels from  the SpaceNet 2 dataset~\cite{van2018spacenet}.
  The first real-world dataset uses actual misaligned data from OpenStreetMap and imagery from the SpaceNet 5 dataset~\cite{van2018spacenet}. The second one, ReBO data~\cite{li2025dragosm} supplies both real misaligned and golden labels. Distributions of building sizes are given in the supplementary materials.}
  \label{tab:dataset_overview}
  \begin{tabular}{lcrrrrcc@{}}
    \toprule
    Cities & Source & Buildings & Mean (m\textsuperscript{2}) & Max (m\textsuperscript{2}) & Patches & Golden & Real\\
    \midrule
    Las Vegas  & SpaceNet 2 & 108,943 & 206 & 28,702   & 13,280 & \ding{51} & \ding{55} \\ % 151,367
    Paris      & SpaceNet 2 &  16,478  & 121 & 14,775   & 2,111  & \ding{51} & \ding{55} \\
    Khartoum   & SpaceNet 2 & 25,113  & 240 & 17,841   & 3,077  & \ding{51} & \ding{55} \\
    San Juan   & SpaceNet 5 \& OSM & 284,042 & 218 & 113,909  & 6,726  & \ding{55} & \ding{51} \\
    41 cities  & ReBO       & 179,265 &  354 & 54,808  & 5473   & \ding{51} & \ding{51} \\
    \bottomrule
  \end{tabular}
\end{table}

The synthetic data was generated for the first three cities in the table. The original image size is $650\times650$ pixels, which we split into patches of $320\times320$ pixels (more details about the data generation are provided in the supplementary material). After applying the systematic bias and uniformly sampled transformations to all image pairs, we split all pairs from each city, including real-world data, into training, validation, and test sets randomly, with $80\%$, $10\%$ and $10\%$, respectively (see \cref{tab:dataset_overview}).

To demonstrate that our method works on  OpenStreetMap~\cite{OpenStreetMap} building data, we collected WorldView-3 imagery from a fourth city, San Juan, through the SpaceNet 5~\cite{van2018spacenet} dataset. These data were released without building footprints, and we downloaded corresponding footprints from OpenStreetMap and prepared them for alignment and segmentation with our methodology. 

In addition, we evaluated our method on ReBO data~\cite{li2025dragosm}, \rebodata{}, a real-world dataset that includes real misaligned and golden labels. The dataset consists of imagery and polygon labels for each building, including footprints, roof, and OSM polygons. The OSM polygons are real-world labels, and the remaining two are manually corrected labels delineating building footprints and roofs, respectively. The imagery consists of both off- and near-nadir images. We use only roof labels as golden labels in our experiments because footprint labels fuse with facade pixels in off-nadir images. While the other datasets were generated for individual cities, \rebodata{} comprises patches from 41 cities worldwide. The dataset was released with a training and test split. We used the training set for both training and validation and reported final results on the test set. The spatial resolution is 50 cm, and each patch in the data has a size of $512\times512$ pixels. See Tab.~\ref{tab:dataset_overview} for more details about this dataset.

\section{Experiments}

In our experiments, the TNet used the ViT-Small~\cite{dosovitskiy2021an} architecture and the SNet was based on a U-Net \cite{ronneberger2015unet}, where the encoder was replaced by ConvNeXt-Tiny~\cite{liu2022convnet}. In all our experiments,  we used  ConvNeXt-Tiny with DINOv3~\cite{simeoni2025dinov3} pretrained weights and froze the encoder. We also experimented with alternative encoders in SNet (see supplementary materials).

We conducted five sets of experiments:
\begin{enumerate}
    \item \textbf{Regularization, augmentation, and end-to-end training.} We evaluated the effects of the transformation bias mitigation strategies. We either trained with $L_\CE$ or $L_\CE+L_\text{Con}$, and either enabled or disabled the large-scale image augmentations, resulting in four different settings. 
    The consistency loss \eqref{eq:loss2} does depend on the SNet output. 
This suggests to pretrain the TNet before training the SNet training, and we evaluated such a two-stage process. 
    All of these experiments were conducted on the Las-Vegas dataset, the two-stage learning was additionally evaluated on \rebodata.

    \item \textbf{Robustness.} To test the robustness of our method against the magnitude of transformations, we varied the magnitude of the applied transformations in \unidata{} and \biasdata{} between 10--100 pixels on the Las-Vegas dataset. For each setting, we re-trained the model and evaluated IoU (see Sec.~\ref{sec:noise_level}).
    
    \item \textbf{Baseline comparison.} We compared our method with five baseline models. These include, two supervised methods, such as MapRepair (MR)~\cite{zorzi2020map}, and Alignment Correction Network (ACN)~\cite{simone2020learning}, and three unsupervised methods, such as Map Alignment (MA)~\cite{Girard_2019}, AutoCorrect (AC)~\cite{chen_autocorrect_2019}, and Spatial Correction (SC)~\cite{yao2023learning}.
    We tried our best to implement these methods for a fair comparison; \eg, we used our TNet architecture in both MR and AC. While some of these baselines were trained to fit only the training set, we used standard data splits such as train, validation, and test sets for all our experiments. 
    SC and MA correct labels by training a model in multiple rounds. SC uses golden labels from the validation dataset and is consequently not a fully unsupervised method. Both MA and our method do not need any golden labels. A crucial hyperparameter of prior work is the number of \emph{rounds} after which the model corrects the labels.
    
    \item \textbf{Real-world dataset.}
    We performed an experiment on the \rebodata{} dataset~\cite{li2025dragosm} that provides real misalignments and golden labels for 41 cities. While they released the patches with the size of $512 \times 512$ pixels, we predicted a single affine transformation for each patch. We used this dataset to train and compare our method with all the baselines.
    
    \item \textbf{Qualitative analysis.} Our method was applied to a real-world dataset with building footprints from OpenStreetMap paired with WorldView-3 imagery.  

\end{enumerate}

\paragraph{Details of the training process.}
In all experiments and for all models, we used the AdamW \cite{adamw} optimizer with a learning rate of $0.00001$ for training. The model was trained for 300 epochs with a batch size of 48. After training, we selected the best model with the weights of the epoch with highest \ioul{} score on the validation set. Since golden labels $\y$ are unavailable in real-world scenarios, \ioua{} and \ioup{} were used only for monitoring and reporting purposes and not for model selection.

\paragraph{Additional hyperparameters.}
We set $c=0.35$ for scaling the $\tanh$ activation, which constrains the translations between $[-112, 112]$ pixels along both axes and the rotation range to be within  $[-20.06, 20.06]$ degrees around the image center. We further set $\lambda = 100$ to scale the two loss terms in \cref{eq:loss2} to a similar order of magnitude.
In the experiments using augmentation, all five geometric augmentations were applied with an equal probability of 50\% to each batch. All experiments were carried out on AMD MI250X (64GB) GPUs.

\begin{table*}[t]
  \centering
  \caption{\textbf{Regularization and augmentations.} We study the effect of the regularization loss $L_{\text{Con}}$ and augmentations. The results show the performance on Las Vegas with the frozen ConvNeXt-Tiny encoder with DINOv3 weights. Both components lead to complementary improvements, with largest improvements on \biasdata.}
  \begin{tabular}{@{}lcccccccc@{}}
    \toprule
    \multicolumn{3}{c}{} & \multicolumn{3}{c}{\unidata}  & \multicolumn{3}{c}{\biasdata} \\
    \cmidrule(lr){4-6} \cmidrule(lr){7-9}
    $L_{\text{CE}}$ & $L_{\text{Con}}$ & Aug. & \ioup & \ioua & \ioul & \ioup & \ioua & \ioul \\
    \midrule
    \checkmark & -           & -          & 0.71 & 0.77 & 0.72 & 0.39 & 0.41 & 0.79 \\
    \checkmark & \checkmark  & -          & 0.74 & 0.79 & 0.72 & 0.41 & 0.43 & 0.78 \\
    \checkmark & -           & \checkmark & 0.66 & 0.69 & 0.79 & 0.72 & 0.76 & 0.82 \\
    \checkmark & \checkmark  & \checkmark & 0.79 & 0.84 & 0.79 & 0.78 & 0.88 & 0.82 \\
    \bottomrule
  \end{tabular}
  \label{tab:aug_results}
\end{table*}
 
\section{Results}

\paragraph{Regularization, augmentation, and end-to-end training.} 
The results for this experiment are given in Tab.~\ref{tab:aug_results}.
The results show that both of our regularization techniques improved \ioup{} and \ioua{} for \unidata{} and \biasdata{} compared to the baseline of only using $L_{\text{CE}}$. Again, the effect was more pronounced for \biasdata, where a sharp decay in both \ioup{} and \ioua{} could be observed in the baseline, while \ioul{} did not vary at all, showing that the SNet was still able to predict the misaligned segmentation maps correctly. In our experiments, adding the augmentations when dealing with the biased \biasdata{} gave the largest improvement, whereas adding only the augmentation to \unidata{} decreased the accuracy.

We compared our end-to-end learning approach with sequential training. We first trained the TNet alone with $L_{\text{Con}}$, then froze it and trained the whole architecture with $L_{\text{CE}}$. This strategy failed to produce the desired results. On \unidata, \biasdata, and \rebodata,
the two-stage training only achieved an \ioup{} of 0.48, 0.44, and 0.38, respectively, and 
an \ioua{} of 0.31, 0.42, and 0.27, respectively (cf.~Tab.~\ref{tab:main_results}).
Learning the whole system with pretrained and frozen TNet seems to make it difficult to provide guiding transformations in early stages of SNet learning. The reason is the TNet pretraining was based on misaligned but otherwise (almost) perfect masks (we added pixel noise) and the SNet produces very noisy masks in the beginning. 
There might be even better ways to combine TNet and SNet optimization than our simultaneous training, but we leave this question to future work.

\begin{table}[ht!]
  \centering
  \caption{\textbf{Comparison to baseline methods.} For each city, the baseline methods MapRepair (MR)~\cite{zorzi2020map}, Alignment Correction Network (ACN)~\cite{simone2020learning},  Spatial Correction (SC)~\cite{yao2023learning}, Map Alignment (MA)~\cite{Girard_2019}, and AutoCorrect (AC)~\cite{chen_autocorrect_2019} are compared to our \emph{AnS} method trained on both \unidata{} and \biasdata{} and evaluated on the respective test set.
  For every city, the first row indicates the IoU metrics of the \emph{initial overlap} of the misaligned labels without any correction. 
  The results on the test set of \rebodata{} are added next to Las Vegas city as additional two columns. (*Denotes supervised baselines.) }
  \label{tab:main_results}\begin{minipage}[t]{.68\textwidth}
  \begin{tabular}[t]{@{}llcccc@{}}
    \toprule
    & & \multicolumn{2}{c}{\unidata}  & \multicolumn{2}{c@{}}{\biasdata}  \\
    \cmidrule(lr){3-4} \cmidrule(lr){5-6} 
    \shortstack{City } & Method & \ioup & \ioua & \ioup & \ioua \\
    \midrule
    Las Vegas   & Initial overlap   & 0.53 & 0.53 & 0.46 & 0.46 \\
    \cmidrule(lr){3-6}
                & MR~\cite{zorzi2020map}*       & --   & 0.77 & --   & 0.77 \\
                & ACN~\cite{simone2020learning}*& 0.94 & --   & 0.93  \\
    \cmidrule(lr){3-6}
                & SC~\cite{yao2023learning}    & 0.66 & --   & 0.47 & --     \\
                & MA~\cite{Girard_2019}        & 0.55 & 0.70 & 0.49 & 0.47  \\
                & AC~\cite{chen_autocorrect_2019}& --   & 0.65 & --   & 0.46  \\
                & AnS (ours)                    & 0.79 & 0.84 & 0.78 & 0.88  \\
     
    \midrule
    Paris       & Initial overlap & 0.47 & 0.47 & 0.40 & 0.40  \\
    \cmidrule(lr){3-6}
                & MR~\cite{zorzi2020map}* &-- & 0.57 & -- & 0.68  \\
                & ACN~\cite{simone2020learning}* & 0.88 & -- & 0.87 & --  \\
    \cmidrule(lr){3-6}
                & SC~\cite{yao2023learning} & 0.48 & -- & 0.40 & --  \\
                & MA~\cite{Girard_2019} & 0.49 & 0.51 & 0.42 & 0.37 \\
                & AC~\cite{chen_autocorrect_2019} & -- & 0.54 & -- & 0.36 \\
                & AnS (ours)        & 0.56 & 0.67 & 0.52 & 0.65 \\
    
    %\cmidrule(lr){1-6}
     \midrule
    Khartoum    & Initial overlap & 0.58 & 0.58 & 0.50 & 0.50  \\
    \cmidrule(lr){3-6}
                & MR~\cite{zorzi2020map}* & -- & 0.71 & -- & 0.75 \\
                & ACN~\cite{simone2020learning}* & 0.91 & -- & 0.90 & --  \\
    \cmidrule(lr){3-6}
                & SC~\cite{yao2023learning}       & 0.60 & -- & 0.47 & --  \\
                & MA~\cite{Girard_2019} & 0.60 & 0.61 & 0.53 & 0.46 \\
                & AC~\cite{chen_autocorrect_2019} & -- & 0.54 & -- & 0.43  \\
                & AnS (ours) & 0.66 & 0.77 & 0.63 & 0.71 \\
    \bottomrule
  \end{tabular}
\end{minipage}\begin{minipage}[t]{.165\textwidth}
\begin{tabular}[t]{@{}cc@{}}
\toprule
\multicolumn{2}{c@{}}{\rebodata} \\
\cmidrule(lr){1-2} 
\ioup & \ioua \\
    \midrule
    0.54 &  0.54 \\
    \cmidrule(lr){1-2}
                    --   & 0.76 \\
                 0.87 & --   \\
                 \cmidrule(lr){1-2} 
                       0.49 & --   \\
               0.55 & 0.57 \\
                 --   & 0.60 \\
                0.62 & 0.74 \\
                \bottomrule
  \end{tabular}
\end{minipage}
\end{table}

\paragraph{Baseline comparison on synthetic and real datasets.} 

The results of our performance comparison are summarized in Tab.~\ref{tab:main_results}.
Our method outperformed all unsupervised baselines (SC, MA, AC) in both \ioup{} and \ioua{}, on both synthetic and real datasets. As expected, both supervised methods (MR, ACN) performed well compared to the unsupervised baselines and our method. Still, our method is competitive with MR, outperforming on \unidata{} and \biasdata{} in Las Vegas, and on \unidata{} in Paris and Khartoum. Note that ACN uses misaligned labels as an input, which means that it required misaligned labels during inference, while no other models require misaligned labels for segmentation. The model performed best on Las-Vegas (which could be an artifact of selecting our architecture on this model), however, it also performed better for Khartoum and Paris, which showed a much smaller initial overlap under our transformations than the other two cities. Note that for Paris, \ioua{} was consistently large for our method, showing that the problem is likely a result of difficulties in predicting the segmentation and a small training dataset, while the transformations were still predicted correctly.

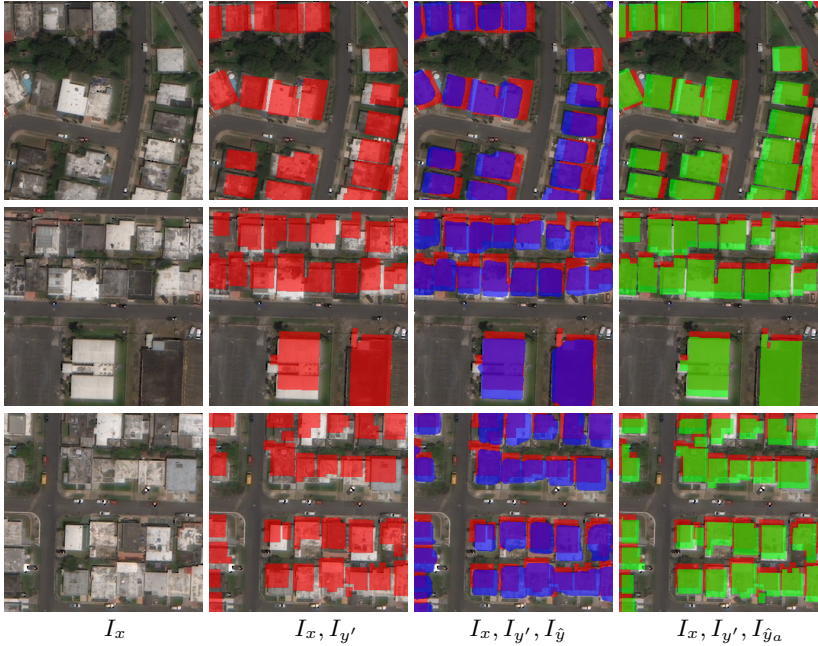
\begin{figure}[t]
    \centering
    \rotatebox{0}{\begin{overpic}[width=0.9\linewidth]{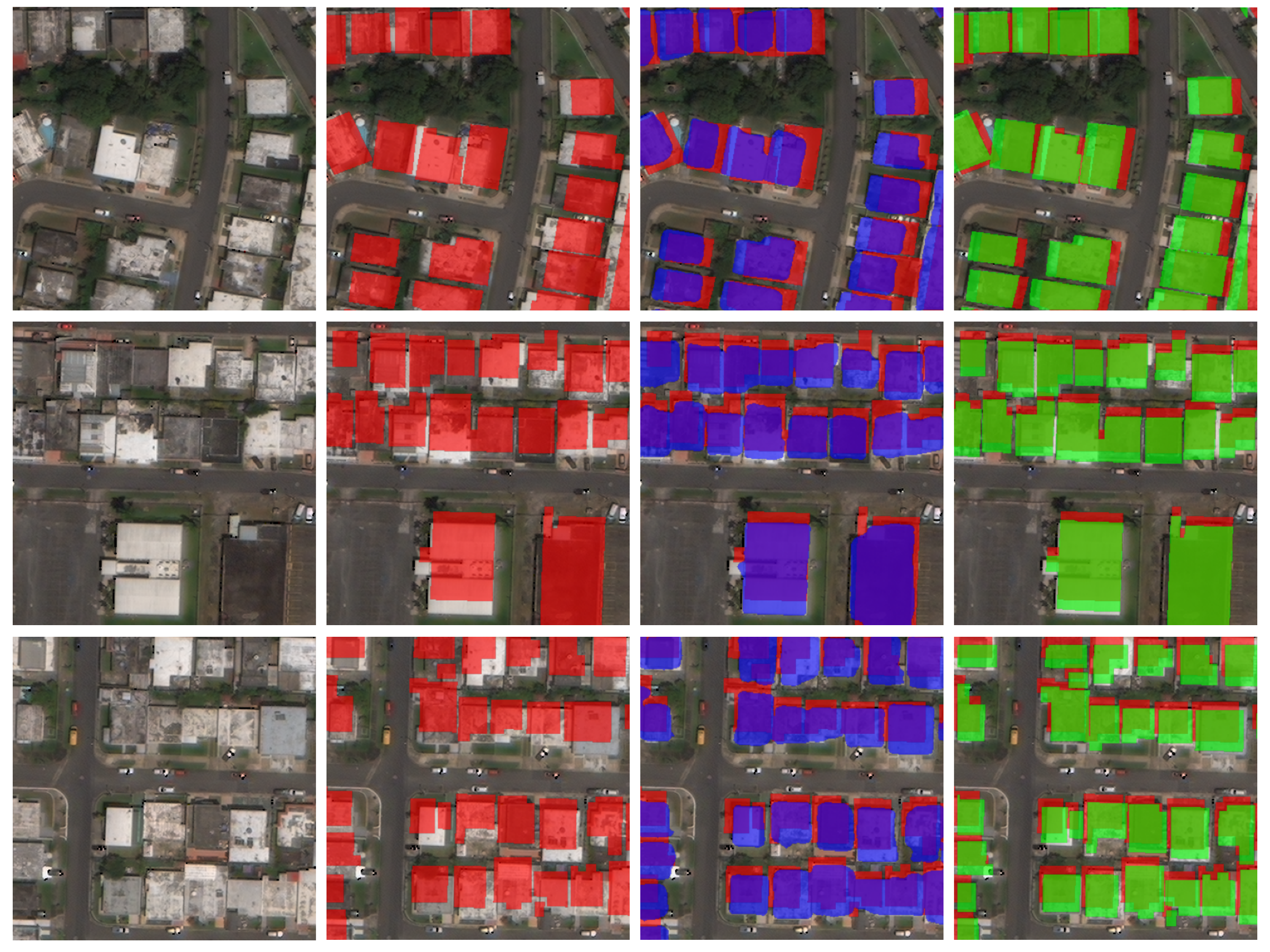}
    \put(12,-2){ $\x$}
    \put(35,-2){ $\x, \yprime$}
    \put(56,-2){ $\x, \yprime, \yhat$}
    \put(81,-2){ $\x, \yprime, \yhata$}
    \end{overpic}}
    \caption{\textbf{Qualitative results learned on OpenStreetMap data.} 
    We show qualitative examples correcting real-world misaligned labels from OpenStreetMap in San Juan using our method. The first column shows the RGB image (\x). All other columns display the \textcolor{red}{misaligned building mask (\yprime)} on top of  \x{} for the reference. The \textcolor{blue}{predicted segmentation mask (\yhat)} and \textcolor{green}{aligned mask (\yhata)} are shown in the third and fourth column, respectively.
    While the output of the semantic segmentation network $\yhat$ has blurry edges and corners (3rd column), the corrected segmentation map $\yhata$ using the estimated affine transformation preserves topology and shapes (last column).
    }
    \label{fig:qualitative_predictions}
\end{figure}

\paragraph{Qualitative analysis on OpenStreetMap data.}
The results of the qualitative analysis on the real application of our method are given in Fig.~\ref{fig:qualitative_predictions}.
Since we do not have golden labels here, we can only estimate the overlap between predicted segmentation map and aligned segmentation map, \ioul{}, and we obtained \ioul=0.70. Three randomly selected examples of the model predictions are given in Fig.~\ref{fig:qualitative_predictions}, and more examples are given in the supplementary material. As can be seen, the predicted transformations correct the misalignment of the label map well. However, the produced label maps are blurry and often do not represent the actual shapes of the houses. This is likely due to the presence of label-errors as we have not included any robust loss components for missing house annotations.

\begin{figure}[t]
    \centering
     \begin{overpic}[width=\linewidth]{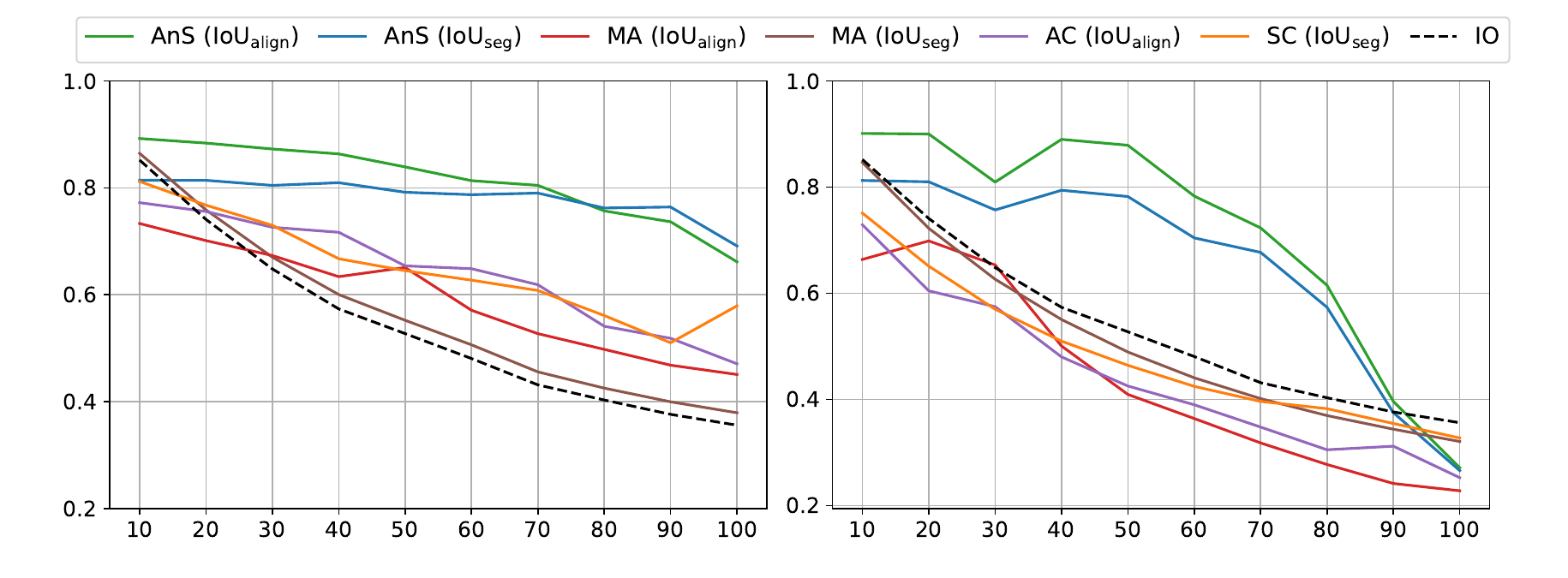}
    \put(12,-0.5){ Random misalignment level}
    \put(56,-0.5){Systematic misalignment level}
    \put(1.5,15){\rotatebox{90} {IoU}}
    \end{overpic}
    \caption{\textbf{Robustness to different noise levels.} Sensitivity analysis of training the model with increasing noise-levels on the random misalignment dataset \unidata{} and systematic misalignment dataset \biasdata{}.} \label{fig:sensitivity}
\end{figure}

\paragraph{Robustness.} \label{sec:noise_level}
Fig.~\ref{fig:sensitivity} shows the results of our robustness experiment with different noise levels. We plot \ioup{} and \ioua{} of our method and all unsupervised baselines, and the IoU of the misaligned transformations with the golden labels~(initial overlap, IO) for comparison.
Compared to the initial overlap, all baselines performed slightly better on \unidata{} but did not improve over the initial overlap on \biasdata{}. The high \ioup{} from our method shows that our model is robust against the magnitude of the applied transformation until 100 pixels in \unidata{} and about 70 pixels in \biasdata{}. Then the \ioup{} starts to drop, which is reflected by a simultaneous drop in \ioua{}.

\section{Discussion}
Our align and segment (AnS) framework has been designed to learn semantic segmentation models from systematically misaligned labels without requiring any access to golden labels. 
On all three cities in the artificial datasets as well as on the real-world data with and without golden labels, our approach demonstrated that jointly estimating label alignment and segmentation is not only feasible but can substantially improve segmentation performance.

Standard segmentation networks readily overfit to misaligned labels. As our experiments show, simply adding a spatial transformer module, referred to as TNet,  does not solve the problem, as it is easily bypassed by the segmentation network. However, our proposed loss term forces the TNet to learn and perform consistent transformations, and the TNet indeed learns to accurately estimate the underlying alignment transformations.
As expected, random misalignments were easier to address than systematic misalignments.
However, we showed that data augmentation in the form of rotating and flipping training images together with their misaligned segmentation masks significantly improves the performance in this setting.

We identified several issues with the baseline methods
considered in our experiments. For example, SC and MA required significant training time due to iterative label correction, incurring computational overhead. In our experiments, we trained these models for at least three rounds as recommended~\cite{yao2023learning, Girard_2019}, which means the computational cost is three times higher than single-round approaches like AC and ours.
MA assigns a single random transformation to the misaligned mask in each round, i.e., the same transformations are used across multiple epochs, whereas our approach samples a new transformation in each epoch.
Resampling transformations every epoch reduces the risk of learning the misalignment.
While AC also uses a similar sampling approach with a consistency loss, there is an important difference: In AC, the transformation network operates on two different modalities, RGB images and segmentation masks, 
whereas our TNet uses two segmentation masks as input.  

\paragraph{Limitations.}
A limitation of our work is the use of a generic decoder for the building footprint segmentation. We neither optimized the decoder for the task at hand nor  used any post-processing.
Specialized methods for building segmentation 
would surely improve the delineation of the building boundaries (\eg, \cite{girard2021polygonal, guthula2025drone}). 

Furthermore, the segmentation is hampered by label noise beyond misalignments, and adding, for example, mechanisms for handling missing labels would improve the training.
The  current model is restricted to learning affine  transformations at the scale of the input patch, but the approach is general enough to allow for more complex transformations.

\section{Conclusion}

Global building footprint data are key for monitoring population densities and managing cities on a large scale. 
Consequently, several organizations now provide large-scale building-footprint datasets, including Google’s Open Buildings~\cite{sirko2021continental} and Microsoft building footprints~\cite{Microsoft}, supplementing globally curated contributions available through OpenStreetMap~\cite{OpenStreetMap}.
Such existing building data could be automatically combined with satellite imagery to construct training datasets to learn semantic segmentation models, which would allow to generate inexpensive high-quality building-footprint data with extensive coverage and high temporal resolution. However, progress is hampered by the misalignment between independent imagery and  existing footprint data.

In this work, we propose \emph{AnS}, a model-agnostic methodology that can be combined with any semantic segmentation model to directly learn the alignment and the segmentation from misaligned labels. We empirically evaluated our end-to-end approach and demonstrated that regularization and data augmentation are key to disentangle segmentation and alignment and to avoid undesired shortcut learning of misaligned labels. On synthetic data, we demonstrate that AnS is robust to high noise levels. Furthermore, we show that our approach can directly learn from real-world misaligned data (e.g., labels from OpenStreetMap) without using any golden labels.

\section*{Acknowledgements}
We acknowledge support through the project 
\emph{Risk-assessment of Vector-borne Diseases Based on Deep Learning and Remote Sensing}  funded by the Novo Nordisk Foundation (grant number NNF21OC0069116).
AK and CI acknowledge additional support by the \emph{Center for Remote Sensing and Deep Learning of Global Tree Resources (TreeSense)} funded by the Danish National Research Foundation (grant number DNRF192). NL and CI acknowledge support by the \emph{Global Wetland Center} (GWC) funded by the Novo Nordisk Foundation (grant number NNF23OC0081089). This work was supported in part by the Pioneer Centre for AI, DNRF grant number P1. We acknowledge the EuroHPC Joint Undertaking for awarding this project access to the EuroHPC supercomputer LUMI, hosted by CSC (Finland) and the LUMI consortium through a EuroHPC Regular Access call.

\bibliographystyle{splncs04}
\bibliography{main}
\clearpage

\appendix
\section{Appendix}

In the following, we provide more details about model architecture (Section~\ref{ape:architecture}),
how we prepared our synthetic datasets (Section~\ref{ape:sythetic_data}) and how we prepared our real OpenStreetMap datasets (Section~\ref{ape:qualitative_data}). We also provided some qualitative and quantitative results in the respective sections.

\subsection{Details of the model architectures} \label{ape:architecture}

\paragraph{\textbf{Encoder of SNet.}} In place of the SNet encoder, we experimented with three different encoders, such as ResNet-34~\cite{resnet},  ConvNeXt-Tiny~\cite{liu2022convnet} and ViT-Small~\cite{dosovitskiy2021an}.
We compared the three encoder backbones and compared random weight initialization to pretrained weights from ImageNet~\cite{ILSVRC15} for ResNet-34 and DINOv3~\cite{simeoni2025dinov3} weights for ConvNeXt-Tiny and ViT-Small. We further evaluated the impact of freezing or fine-tuning the initialized weights in encoders during training (See Tab.~\ref{tab:freez_results}).

\begin{table*}[b!]
  \centering
  \caption{\textbf{Freezing vs.~training the segmentation encoders.}  Results for different encoders used in the SNet. We experiment with different pretrained encoders as well as with different training strategies that either freeze or train (finetune) the encoders jointly with the TNet. These are our initial experimental results that were trained when we used a $c$ parameter of 0.2 for the output prediction range.}
  \begin{tabular}{@{}lcccccccccc@{}}
    \toprule
    & & \multicolumn{4}{c}{\unidata}  & \multicolumn{4}{c}{\biasdata} \\
    \cmidrule(lr){3-6} \cmidrule(lr){7-10}
    & & \multicolumn{2}{c}{Freeze}  & \multicolumn{2}{c}{Train} & \multicolumn{2}{c}{Freeze}  & \multicolumn{2}{c}{Train} \\
    \cmidrule(lr){3-4} \cmidrule(lr){5-6} \cmidrule(lr){7-8} \cmidrule(lr){9-10}
    \shortstack{Encoder} & \shortstack{Weights} & \ioup & \ioua & \ioup & \ioua & \ioup & \ioua & \ioup & \ioua \\
    \midrule
    ResNet-34  & Random    & 0.60 & 0.72 & 0.68 & 0.71 & 0.61 & 0.75 & 0.71 & 0.77 \\
    ResNet-34  & ImageNet  & 0.80 & 0.86 & 0.78 & 0.80 & 0.68 & 0.72 & 0.41 & 0.43 \\
    ConvNeXt   & DINOv3    & 0.81 & 0.85 & 0.72 & 0.72 & 0.69 & 0.72 & 0.41 & 0.43 \\
    ViT-Small  & DINOv3    & 0.79 & 0.84 & 0.72 & 0.73 & 0.70 & 0.73 & 0.40 & 0.41 \\
    \bottomrule
  \end{tabular}
  \label{tab:freez_results}
\end{table*}

The results of our comparison of model architectures and initialization choices are shown in Tab.~\ref{tab:freez_results}. For all weight initializations (except random) and all datasets, freezing the pre-initialized weights outperformed fine-tuning. Indeed, we can see that for the frozen weights, the networks showed reduced bias, which is reflected in the larger \ioup{} and \ioua{} compared to trained variants. This effect was more pronounced on \biasdata{} than on \unidata. Finally, pre-trained initializations consistently outperformed random initialization in this study; overall, the ConvNeXt backbone with pretrained weights from DINOv3 performed best.

\paragraph{\textbf{Decoder of SNet.}}
When using ResNet-34 as an encoder, we used the standard U-Net decoder. The decoder for ConvNeXt-Tiny and ViT-Small was composed of convolutional and bilinear upsampling layers~\cite{ronneberger2015unet}, where each convolution was followed by a GeLU activation function~\cite{hendrycks2016gelu}. 
Based on the backbone, we adapted the number of skip connections from the different encoder stages. All variants employed skip connections at the input layer at full resolution, implemented using two convolutional layers of kernel size $1\times1$. The ResNet-34, ConvNeXt-Tiny and ViT-Small models used additional three, two, and no skip-connections, respectively, each with a convolution using a $3\times 3$ kernel. 

\subsection{Synthetic Dataset} \label{ape:sythetic_data}

\subsubsection{Data preparation.}
\begin{figure}[ht]
    \centering
    \includegraphics[width=0.75\linewidth]{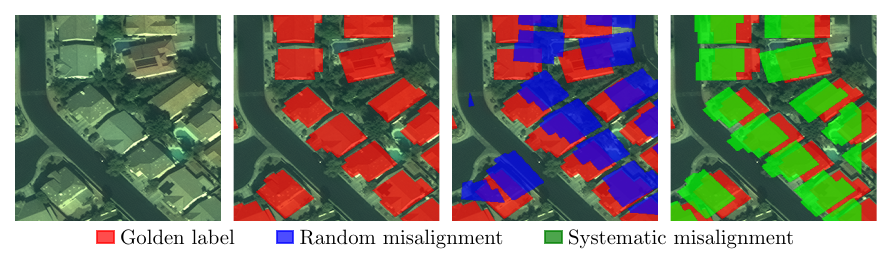}
    
    \caption{ Examples of \x{} (input image), \y{} (golden label) and \yprime{} (artificially misaligned labels). In the third column, \yprime{} was generated by assigning random misalignment with magnitude of $\|\Atheta-Id\|_F =0.241$, $t_x=-48$, $t_y=49$, and $r \approx 4.5^{\circ}$.
    In the fourth column, \yprime{} was generated using systematic misalignment with magnitude of $\|\Atheta-Id\|_F =0.156$, $t_x=50$, $t_y=0$, and $r \approx 0^{\circ}$.
    }
    \label{fig:patch_example}
\end{figure}

We created misaligned datasets for three cities in the SpaceNet-2~\cite{van2018spacenet} benchmark by applying affine transformations to the labels in every patch. We created patches of $320 \times 320$ pixels from the original $650\times650$ patches without any overlap to predict misalignment for a smaller region. To generate the dataset \biasdata{}, we apply the same misalignment for every patch, i.e., 50 pixels translation on the x-axis (see last image in Fig~\ref{fig:patch_example}). We used this transformation in all but the robustness experiment.

\newcommand{\smax}{s_{\max}}
To generate the random dataset \unidata{}, we assigned a random misalignment to every patch. This random misalignment depends on a single parameter $\smax$ limiting the \emph{maximum shift}.
When we generate random misalignment, we need the translation parameters $t_x$, $t_y$ in pixels and rotation angle $\alpha$ in radians to create an affine transformation \Atheta.
We sample $N$ sets of parameters, by sampling $t_x,t_y$ uniformly in $[\smax, \smax]$ and $\alpha$ uniformly in $[-\frac {\smax}{2\cdot320},\frac {\smax}{2\cdot320}]$. As a result, with $\smax = 50$, the range of the resulting rotations is $[-4.5, 4.5]$ degrees.
See Fig.~\ref{fig:patch_example} for an example from \unidata{} and \biasdata{}.

\begin{figure}
\centering
    \includegraphics[width=0.60\linewidth]{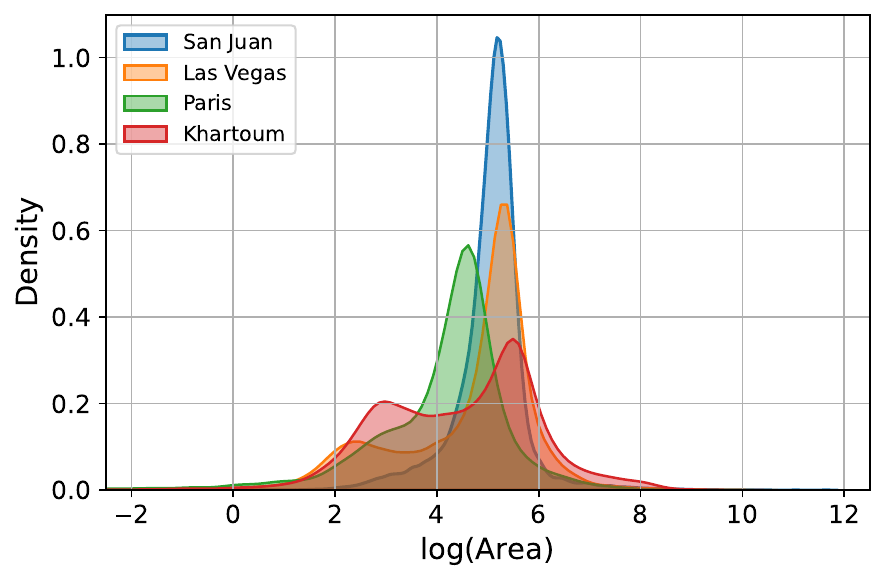}
    \caption{\textbf{Distribution of area of buildings in all cities.} The area of buildings is measured in square meters before applying a log scale on the x-axis. The size of buildings varies across cities. For example, it is high in Las Vegas and San Juan and concentrated, and it is smaller in Paris compared to Las Vegas and San Juan. Buildings in Khartoum are distributed with two peaks. The size of buildings in all these cities can be observed in figures that show predictions (from Fig~\ref{fig:vegas_u} to Fig~\ref{fig:khartoum_b}). 
    } \label{fig:data_distribution}
\end{figure}

\subsubsection{Predictions.}
\begin{figure}
\centering
    \includegraphics[width=1.0\linewidth]{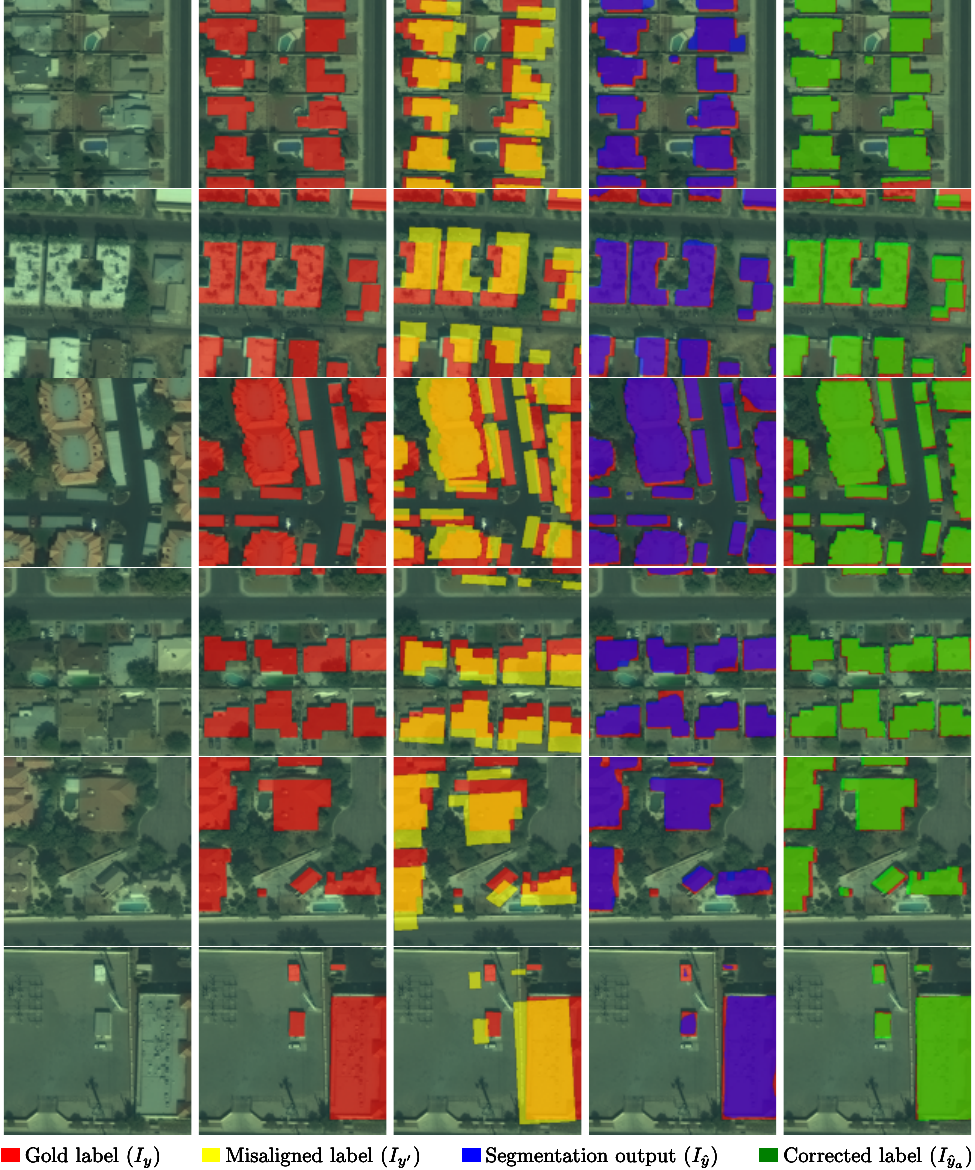}
    \caption{\textbf{Example images and predictions from \unidata{} of Las Vegas}. The first column shows the RGB image (\x) and the second column shows the golden labels (\y). The third column presents the misalignment by comparing the misaligned labels (\yprime) with $\y$. The fourth and fifth columns present predicted segmentation output (\yhat) and corrected label (\yhata) quality by comparing both with \y.
    } \label{fig:vegas_u}
\end{figure}

\begin{figure}
\centering
    \includegraphics[width=1.0\linewidth]{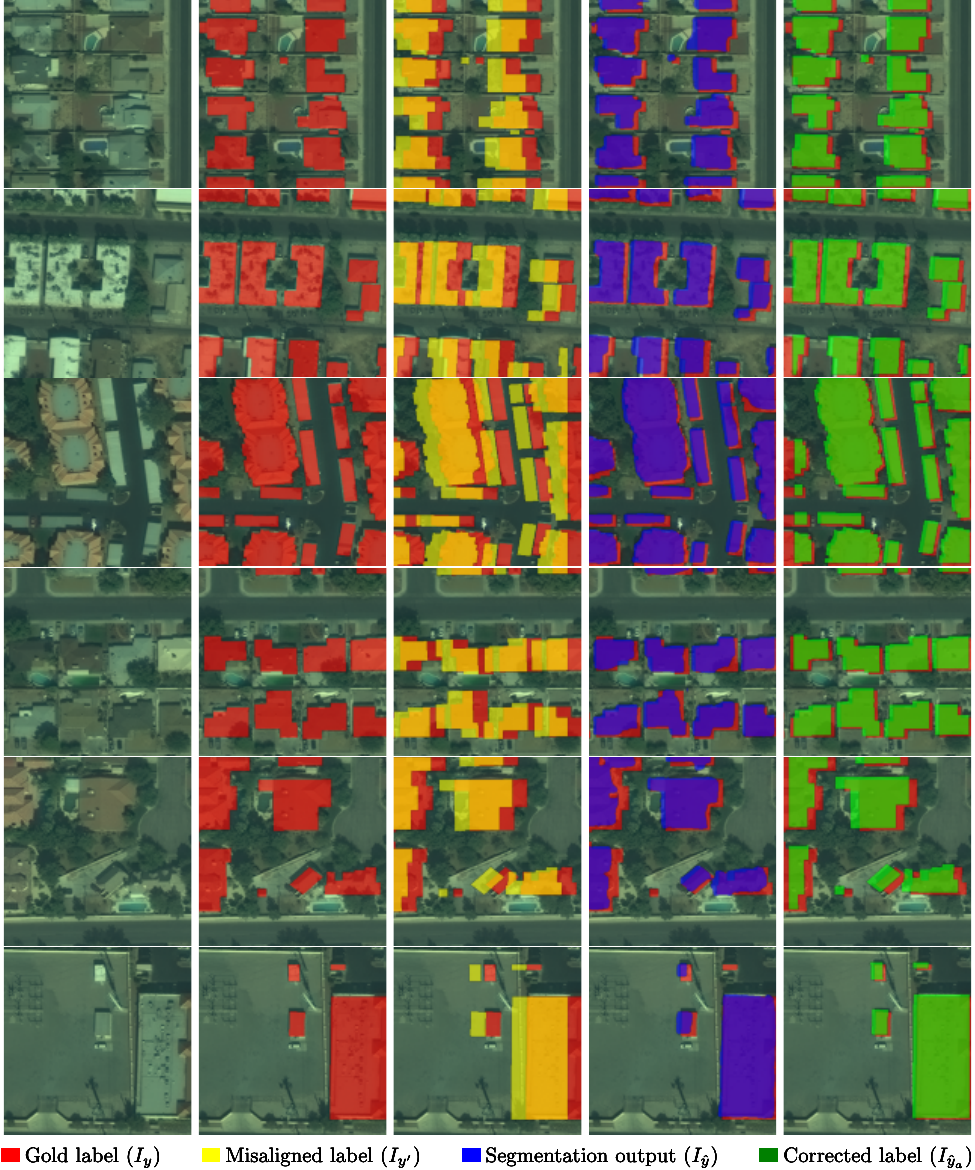}
    \caption{\textbf{Example images and predictions from \biasdata{} of Las Vegas.}
    The first column shows the RGB image (\x) and the second column shows the golden labels (\y). The third column presents the misalignment by comparing the misaligned labels (\yprime) with $\y$. The fourth and fifth columns present predicted segmentation output (\yhat) and corrected label (\yhata) quality by comparing both with \y.
    } \label{fig:vegas_b}
\end{figure}

\begin{figure}
\centering
    \includegraphics[width=1.0\linewidth]{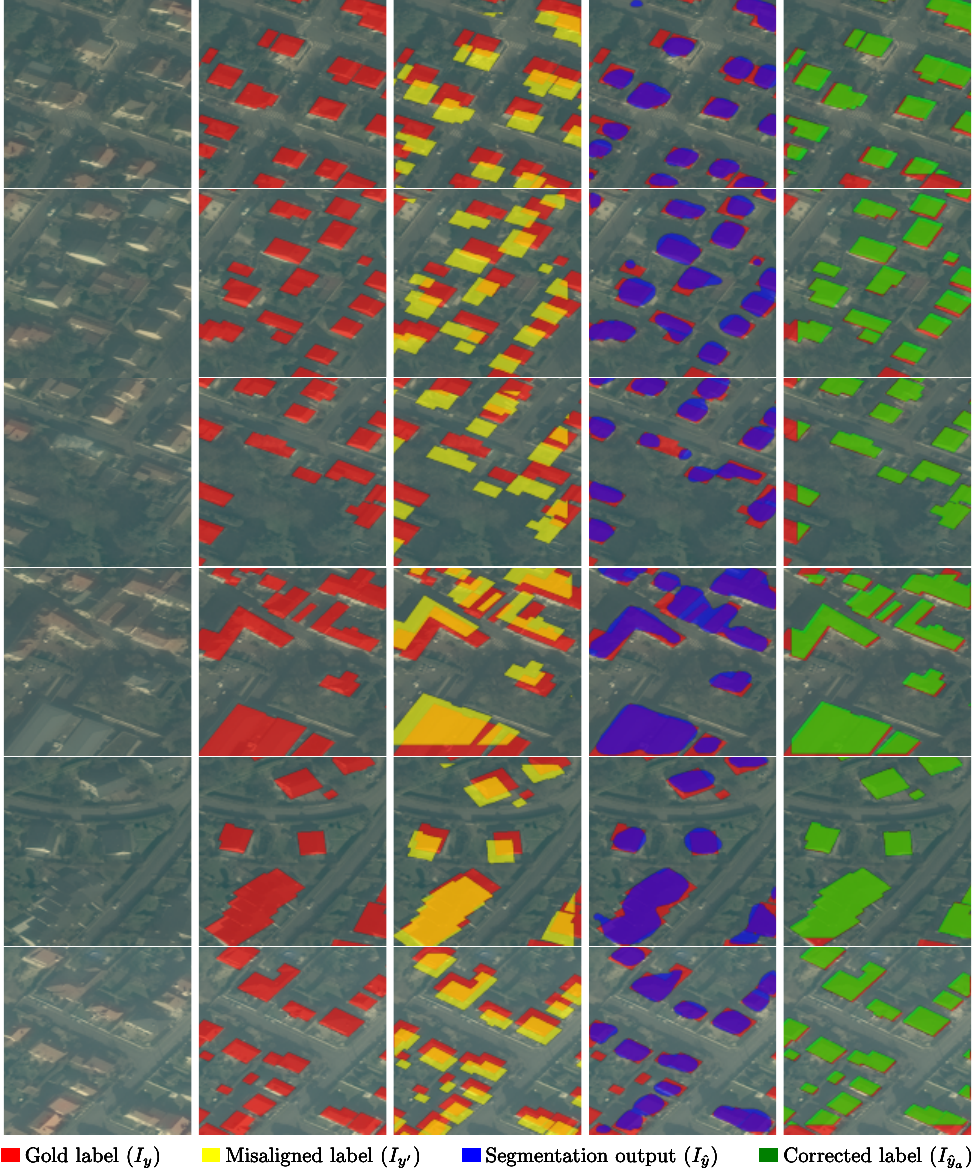}
    \caption{\textbf{Example images and predictions from \unidata{} of Paris}
    The first column shows the RGB image (\x) and the second column shows the golden labels (\y). The third column presents the misalignment by comparing the misaligned labels (\yprime) with $\y$. The fourth and fifth columns present predicted segmentation output (\yhat) and corrected label (\yhata) quality by comparing both with \y.
    } \label{fig:paris_u}
\end{figure}

\begin{figure}
\centering
    \includegraphics[width=1.0\linewidth]{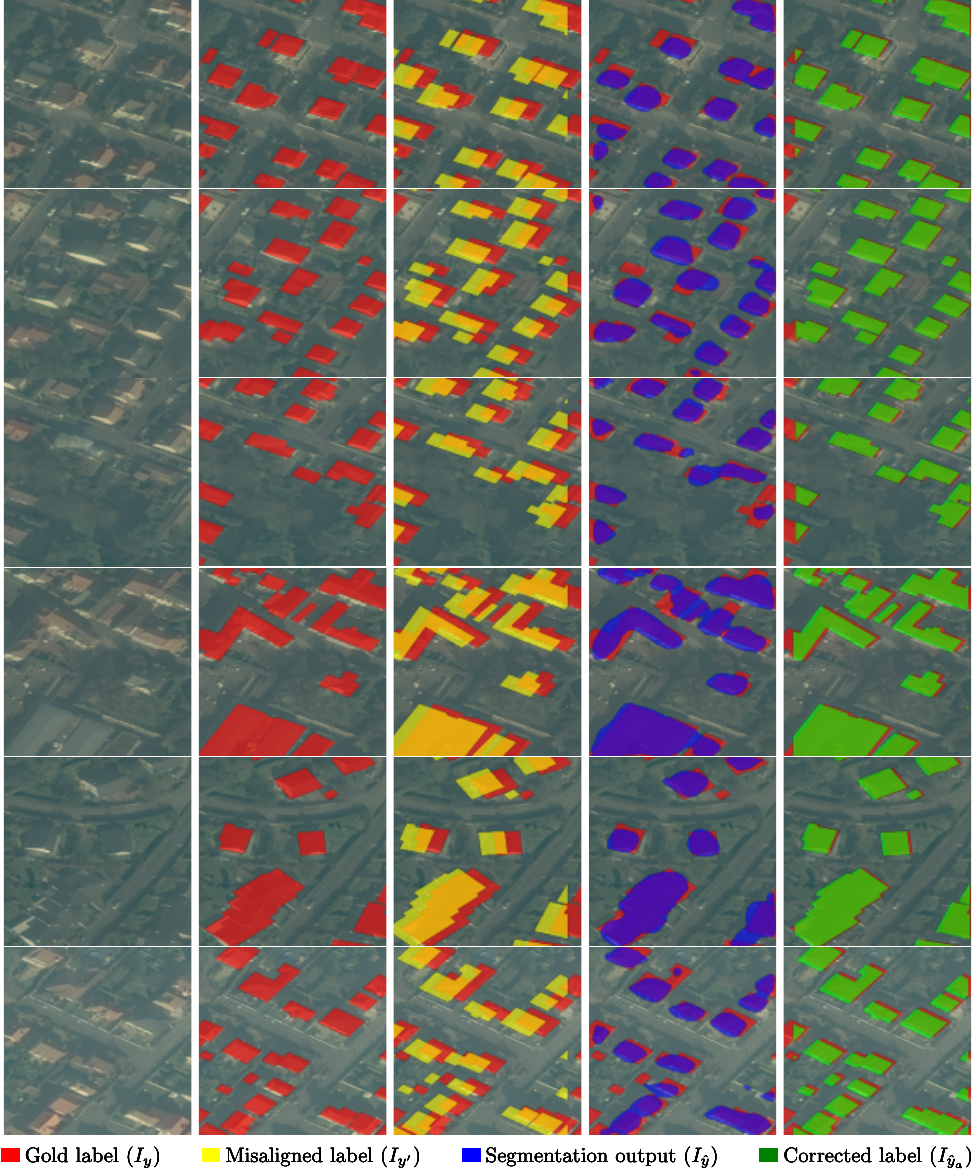}
    \caption{\textbf{Example images and predictions from \biasdata{} of Paris}.
    The first column shows the RGB image (\x) and the second column shows the golden labels (\y). The third column presents the misalignment by comparing the misaligned labels (\yprime) with $\y$. The fourth and fifth columns present predicted segmentation output (\yhat) and corrected label (\yhata) quality by comparing both with \y.
    } \label{fig:paris_b}
\end{figure}

\begin{figure}
\centering
    \includegraphics[width=1.0\linewidth]{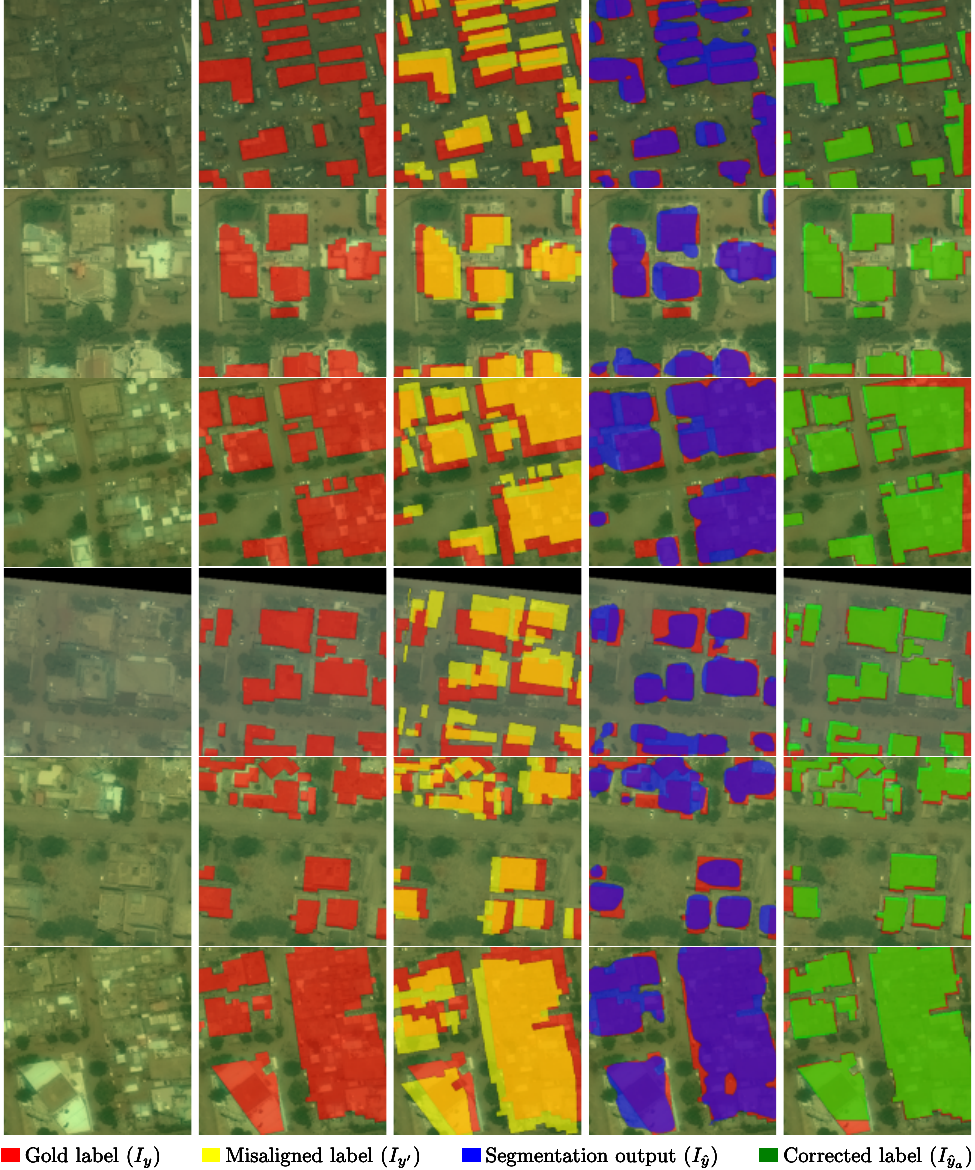}
    \caption{\textbf{Example images and predictions from \unidata{} of Khartoum.}
    The first column shows the RGB image (\x) and the second column shows the golden labels (\y). The third column presents the misalignment by comparing the misaligned labels (\yprime) with $\y$. The fourth and fifth columns present predicted segmentation output (\yhat) and corrected label (\yhata) quality by comparing both with \y.
    } \label{fig:khartoum_u}
\end{figure}

\begin{figure}
\centering
    \includegraphics[width=1.0\linewidth]{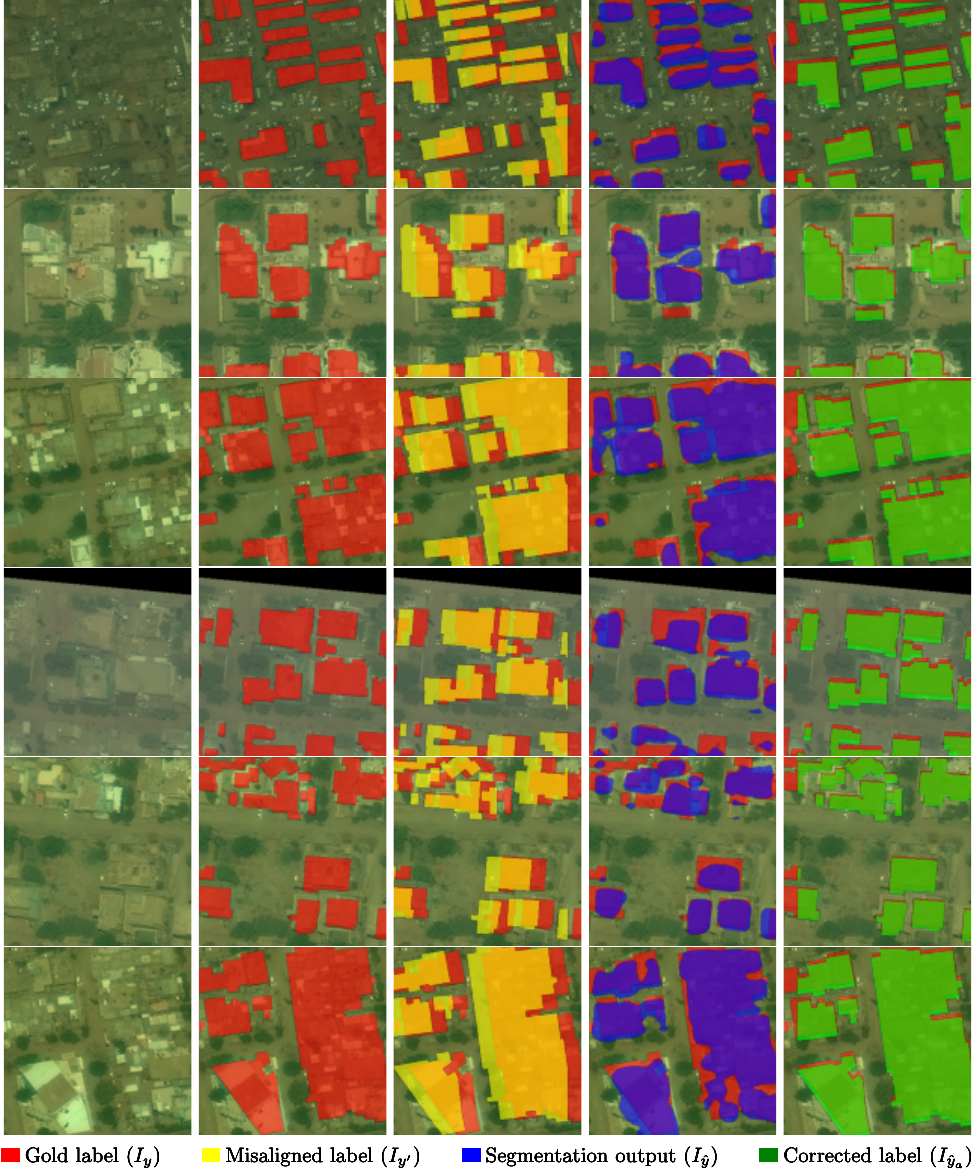}
    \caption{\textbf{Example images and predictions from \biasdata{} of Khartoum.}
    The first column shows the RGB image (\x) and the second column shows the golden labels (\y). The third column presents the misalignment by comparing the misaligned labels (\yprime) with $\y$. The fourth and fifth columns present predicted segmentation output (\yhat) and corrected label (\yhata) quality by comparing both with \y.
    } \label{fig:khartoum_b}
\end{figure}

Since the alignment and segmentation performance also depends on the size of buildings, the distribution of building size is visualized in Fig.~\ref{fig:data_distribution}.
Compared to other cities, the size of buildings is bigger in Las Vegas, and this leads to good overlap between original buildings (or golden labels) and misaligned labels. This can be a good reason why the performance in this city is better compared to the other two cities. While the alignment masks corrected very well with better \ioua{}, the segmentation masks required some improvements. 
We presented some examples of images and predictions in each city. 
Fig.~\ref{fig:vegas_u} and Fig.~\ref{fig:vegas_b} shows examples from \unidata{} and \biasdata{}, respectively, from Las Vegas.
Fig.~\ref{fig:paris_u} and Fig.~\ref{fig:paris_b} shows examples from \unidata{} and \biasdata{}, respectively, from Paris. 
The buildings are smaller compared to Las Vegas, leading to less or no overlap between golden labels and misaligned masks. This can be the reason for low scores of both \ioup{} and \ioua{}. Fig.~\ref{fig:khartoum_u} and Fig.~\ref{fig:khartoum_b} shows examples from \unidata{} and \biasdata{}, respectively, from Khartoum city. The sizes of buildings in the city range from small to large.

\subsection{OpenStreetMap dataset} \label{ape:qualitative_data}
\subsubsection{Data preparation.}
\begin{figure}[ht!]
\centering
    \includegraphics[width=1.0\linewidth]{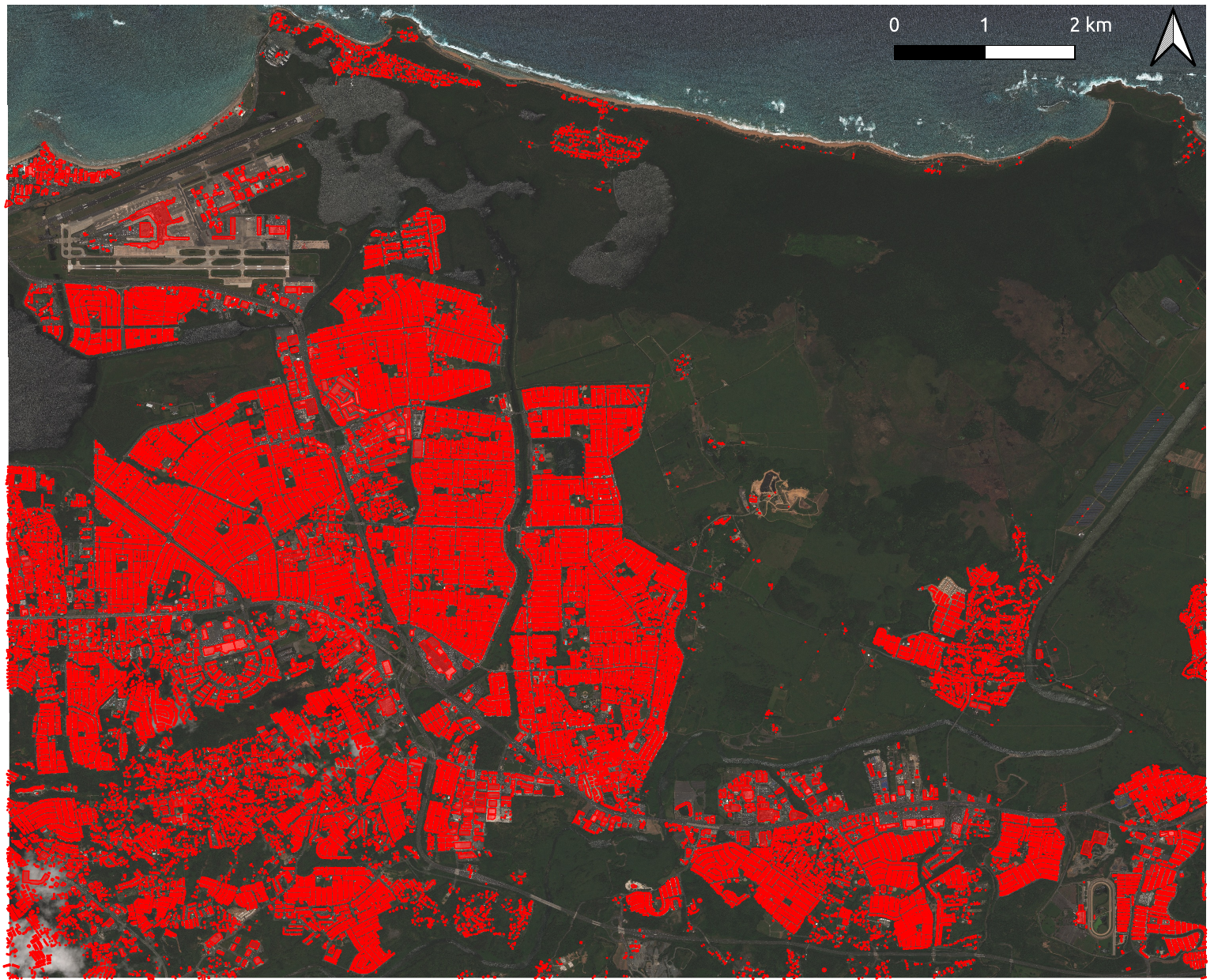}
    \caption{\textbf{Imagery and OpenStreetMap building footprints of San Juan city.}
    } \label{fig:city_buildings}
\end{figure}

Fig.~\ref{fig:city_buildings} shows the imagery and building footprints we used as a part of qualitative analysis on a real OpenStreetMap dataset. We obtained original imagery from SpaceNet-5~\cite{van2018spacenet} and building footprints from OpenStreetMap. Then the entire imagery divided into small patches of $320\times320$ pixels without any overlap. Patches without building footprints are removed, and the remaining patches are split into training, validation, and test sets in an 8:1:1 ratio.

\subsubsection{Predictions.}
\begin{figure}
\centering
    \includegraphics[width=1.0\linewidth]{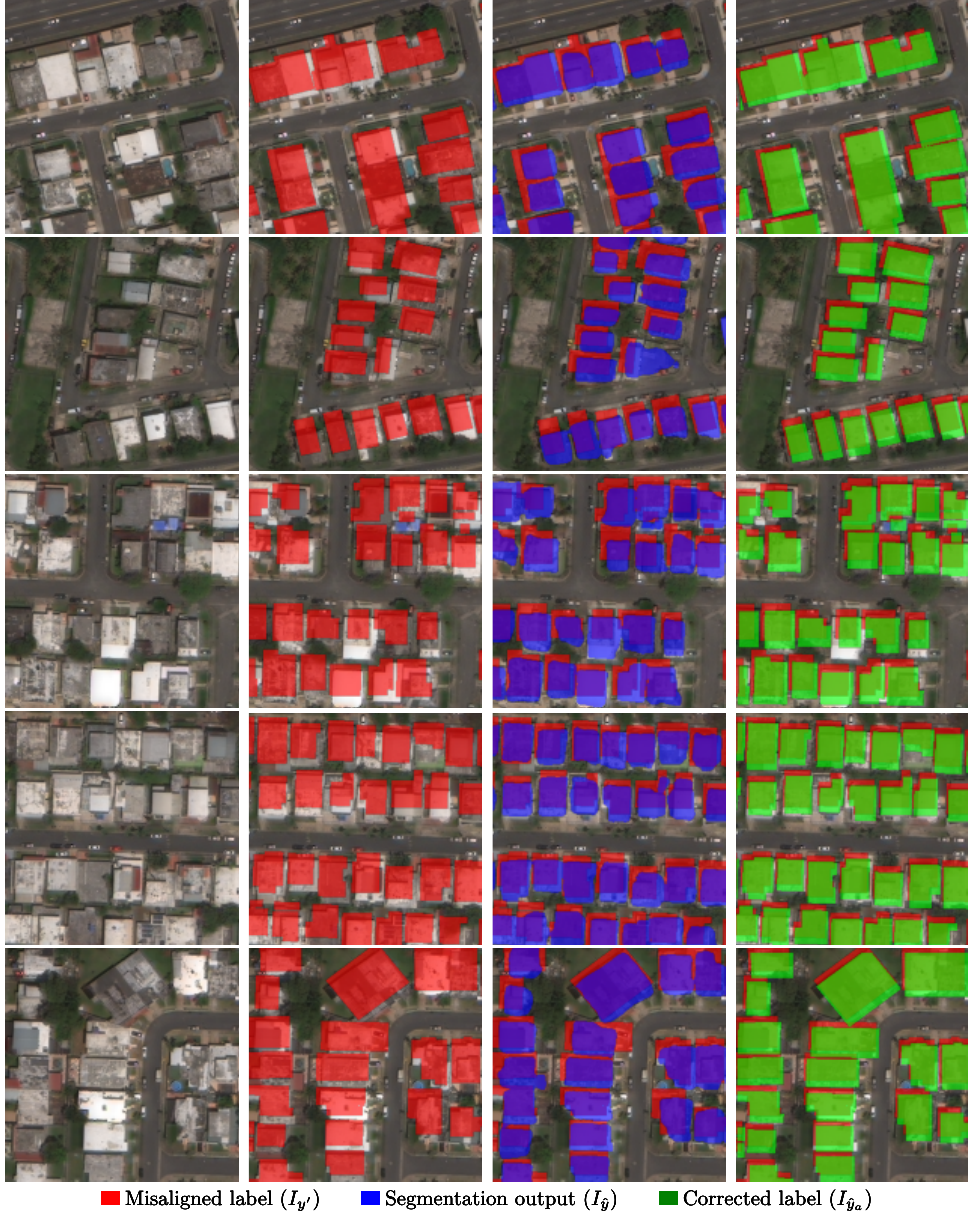}
    \caption{\textbf{Example images and predictions from San Juan city data.}
    The first column shows the RGB image (\x) and the second column shows the misaligned labels (\yprime) from OpenStreetMap data. The third and fourth columns present predicted segmentation output (\yhat) and corrected label (\yhata) quality by comparing both with \yprime.
    } \label{fig:sanjuan_r}
\end{figure}

\begin{figure}
\centering
    \includegraphics[width=1.0\linewidth]{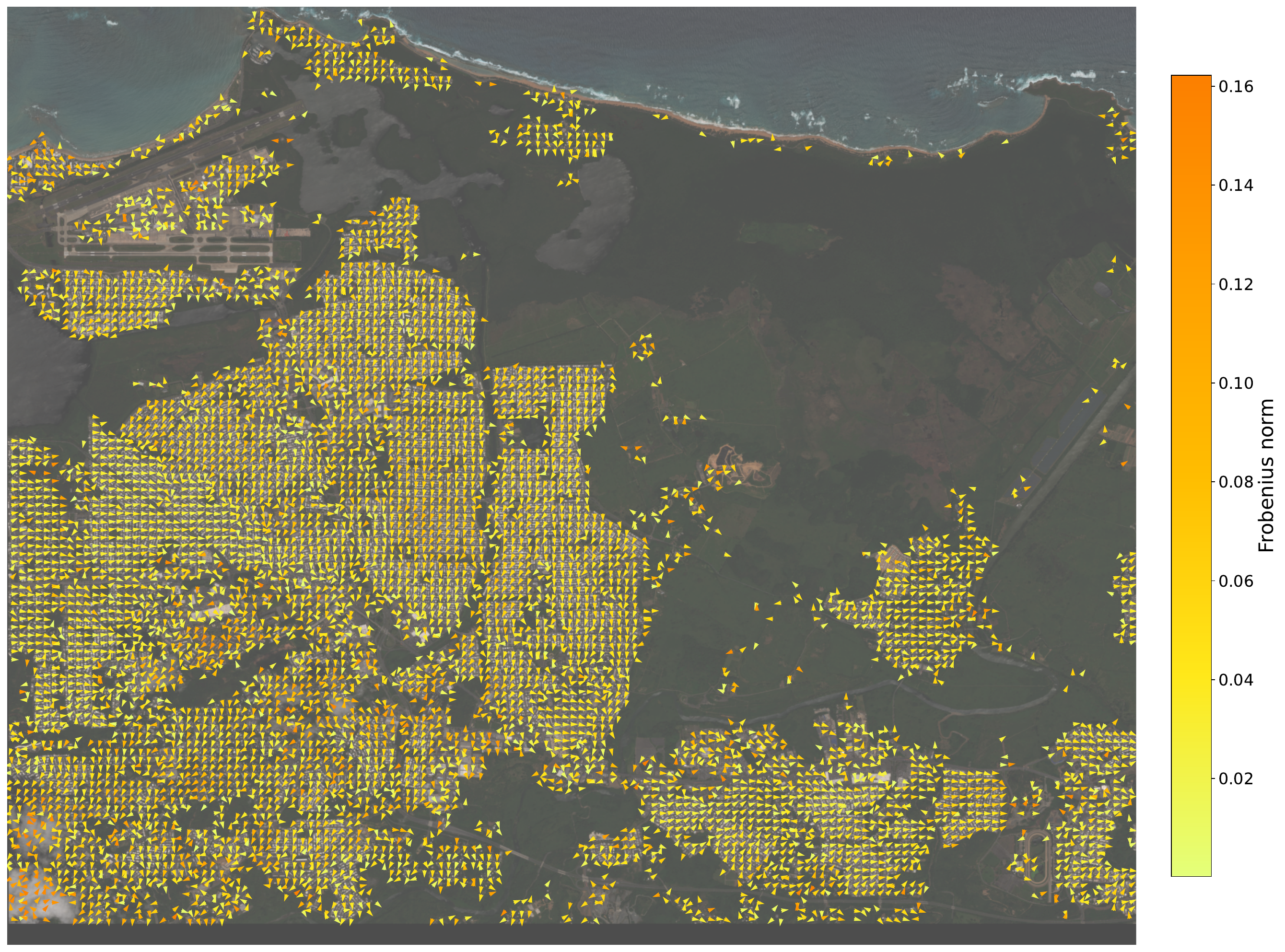}
    \caption{\textbf{The predicted transformation over the entire San Juan city.} The flow field was generated using the translations predicted using the model. The colorbar shows the magnitude of the predicted transformation.
    } \label{fig:flow_field}
\end{figure}

Fig.~\ref{fig:sanjuan_r} shows more qualitative examples of predictions and Fig.~\ref{fig:flow_field} shows the translation of the predicted affine transformation of every patch as a flow field. 
The flow field only prepared using translations $t_x$ and $t_y$ because of predictions of a very small rotation. 
The color represents the magnitude of the predicted transformation that estimated the Frobenius norm of \Athetai{} after subtracting an identity matrix from it.

\end{document}